\documentclass[letterpaper]{article} 
\usepackage{aaai24}  
\usepackage{times}  
\usepackage{helvet}  
\usepackage{courier}  
\usepackage[hyphens]{url}  
\usepackage{graphicx} 
\usepackage{natbib}  
\usepackage{caption} 
\usepackage{subcaption}
\usepackage{algorithm}
\usepackage{amsmath}
\usepackage{amssymb}
\usepackage{multirow}
\usepackage{booktabs}
\usepackage{subcaption}
\usepackage{algpseudocode}
\newcommand{\ie}{\textit{i}.\textit{e}., }
\usepackage{color} 

\DeclareMathOperator*{\argmax}{arg\,max}

\DeclareMathOperator{\sign}{sign}

\usepackage{newfloat}
\usepackage{listings}
\DeclareCaptionStyle{ruled}{labelfont=normalfont,labelsep=colon,strut=off} 
\floatstyle{ruled}
\newfloat{listing}{tb}{lst}{}
\floatname{listing}{Listing}
\title{Cross-Class Feature Augmentation for Class Incremental Learning}
\author{$\text{Taehoon Kim}^{1}$ \quad\quad\quad \; $\text{Jaeyoo Park}^{1}$ \quad\quad\quad \; $\text{Bohyung Han}^{1,2}$
}
\affiliations{
   $\text{ECE}^{1}$ \& $\text{IPAI}^{2}$, Seoul National University\\
    {\tt\small \{kthone, bellos1203, bhhan\}@snu.ac.kr}
}

\usepackage{bibentry}

\begin{document}

\maketitle


\begin{abstract}
We propose a novel class incremental learning approach, which incorporates a feature augmentation technique motivated by adversarial attacks.
We employ a classifier learned in the past to complement training examples of previous tasks. 
The proposed approach has an unique perspective to utilize the previous knowledge in class incremental learning since it augments features of arbitrary target classes using examples in other classes via adversarial attacks on a previously learned classifier.
By allowing the Cross-Class Feature Augmentations (CCFA), each class in the old tasks conveniently populates samples in the feature space, which alleviates the collapse of the decision boundaries caused by sample deficiency for the previous tasks, especially when the number of stored exemplars is small.
This idea can be easily incorporated into existing class incremental learning algorithms without any architecture modification. 
Extensive experiments on the standard benchmarks show that our method consistently outperforms existing class incremental learning methods by significant margins in various scenarios, especially under an environment with an extremely limited memory budget.
\end{abstract}


\section{Introduction}
\label{sec:intro}

Recent deep learning techniques have shown remarkable progress in various computer vision tasks including image classification~\cite{he2016deep,hu2018squeeze}, object detection~\cite{liu2016ssd,redmon2016you,zhu2020deformable}, semantic segmentation~\cite{chen2017deeplab,long2015fully,noh2015learning}, and many others.
Behind this success is an implicit assumption that the whole dataset with a predefined set of classes should be given in a batch.
However, this assumption is unlikely to hold in the real-world scenarios which change dynamically over time. 
This limits the applicability to real-world problems because deep neural networks trained under changing data distribution often suffer from catastrophic forgetting, meaning that the models lose the ability to maintain knowledge about old tasks.
While a straightforward way to handle the critical challenge is to retraining the model with an integrated dataset, this is too expensive or even impossible due to the limitation of computational resources and the inaccessibility of training data.

Class incremental learning is a framework that progressively increases the number of classes while combating the catastrophic forgetting issue.
Among many existing approaches~\cite{aljundi2018memory,kirkpatrick2017overcoming,li2017learning,rebuffi2017icarl,wu2019large,zenke2017continual}, the techniques based on knowledge distillation with exemplars~\cite{douillard2020podnet,hou2019learning,li2017learning}, allowing new models to mimic previous ones, have demonstrated promising performance in alleviating the feature drift issue.
Yet, these methods still suffer from data deficiency for old tasks and data imbalance between tasks, as only few training examples are available for the previous tasks.
To alleviate these limitations, some existing approaches generate either data samples~\cite{ostapenko2019learning,shin2017continual} or feature representations~\cite{liu2020generative}.
However, they require additional generative models, hampering the stability of convergence and increases the complexity of models.

This paper presents a novel feature augmentation technique, referred to as Cross-Class Feature Augmentation (CCFA), which effectively tackles the aforementioned limitations in class incremental learning. 
By leveraging the representations learned in the past, we aim to augment the features at each incremental stage to address data deficiency in the classes belonging to old tasks.
To this end, inspired by adversarial attacks, we adjust the feature representations of training examples to resemble representations from specific target classes that are different from their original classes.
These perturbed features allow a new classifier to maintain the decision boundaries for the classes learned up to the previous stages. 
Note that this is a novel perspective different from conventional adversarial attack methods~\cite{carlini2017cw,goodfellow2015fgsm,madry2018pgd,sey2016df,zhao2018nat}, which focus on deceiving models. 
One may consider generating additional features for each class using the exemplars with the same class labels.
However, this strategy is prone to generate redundant or less effective features for defending class boundaries, especially when the number of exemplars in each class is small.
On the contrary, the proposed approach exploits exemplars in various classes observed in the previous tasks as well as a large number of training data in the current task, which is helpful for synthesizing features with heterogeneous properties.

%
The contributions of this paper are summarized below:
\begin{itemize}
	
	\item[$\bullet$] We propose a novel class incremental learning technique, which effectively increases training examples for old tasks via feature augmentation.
	 The proposed method prevents catastrophic forgetting without modifying architectures or introducing generative models. \vspace{0.1cm}
	

	\item[$\bullet$] Our Cross-Class Feature Augmentation (CCFA) synthesizes augmented features across class labels using a concrete objective motivated by adversarial attacks. \vspace{0.1cm}
			
	\item[$\bullet$] Our algorithm is easily incorporated into existing class incremental learning methods and improves performance consistently on multiple datasets with diverse scenarios, especially under minimal memory budgets.
\end{itemize}

\section{Related Works}
\label{sec:related}

This section reviews existing algorithms related to class incremental learning and adversarial attacks. 

\subsection{Class Incremental Learning}
Most of the existing class incremental learning algorithms address the catastrophic forgetting issue using the following techniques: 1) parameter regularization, 2) architecture expansion, 3) bias correction, 4) knowledge distillation, and 5) rehearsal.

\paragraph{Parameter regularization}
The methods that belong to this category~\cite{aljundi2018memory,kirkpatrick2017overcoming,zenke2017continual} measure the importance of each model parameter and determine its flexibility based on its importance.
The popular metrics to determine the plasticity of models on new tasks include the Fisher information matrix~\cite{kirkpatrick2017overcoming}, the path integral along parameter trajectory~\cite{zenke2017continual}, and the output vector changes~\cite{aljundi2018memory}.
However, their empirical generalization performances are not satisfactory in class incremental learning scenarios~\cite{hsu2018re,van2019three}.

\paragraph{Architecture expansion}
Architectural methods~\cite{rusu2016progressive,yoon2017lifelong} typically focus on expanding network capacity dynamically to handle a sequence of incoming tasks.
To this end, \cite{yan2021dynamically}~proposes to expand the network for each incoming task while compressing the network after the end of each task for computational efficiency.
Recently, \cite{liu2021adaptive}~adopts two network blocks to balance plasticity and stability, which are optimized by a bi-level optimization.
Although these approaches show remarkable performances even with a long sequence of tasks, their computational complexity at both training and inference time increases linearly as the number of tasks grows, which leads to a significant computational burden and challenges in inference.

\paragraph{Bias correction}
There exist several approaches~\cite{hou2019learning,wu2019large} that tackle the bias towards new classes incurred by class imbalance. 
To be specific, \cite{wu2019large} reduces the bias by introducing additional scale and shift parameters for an affine transformation of the logits for new classes.
Moreover, \cite{zhao2020maintaining} matches the scale of the weight vectors for the new classes with the average norm of the old weight vectors.

\paragraph{Knowledge distillation}
The methods based on knowledge distillation~\cite{hinton2015distilling,romero2014fitnets,zagoruyko2016paying,park2021class,kang2022class} aim to encourage a model to learn new tasks while mimicking the representations of the old model trained for the previous tasks.
To this end,~\cite{li2017learning,rebuffi2017icarl,wu2019large,castro2018end,hou2019learning} match their outputs after the classification layers with old models to preserve the representations of input examples.
PODNet~\cite{douillard2020podnet} controls the balance between the previous knowledge and the new information by preserving the relaxed representations.
Here, relaxed representations are obtained by applying the sum pooling along the width and height dimensions to the original intermediate feature maps.
Recently, \cite{simon2021learning}~proposes to optimize the model considering the concept of the geodesic flow and AFC~\cite{kang2022class} introduces the way to handle catastrophic forgetting by minimizing the upper bound of the loss increases caused by the representation change.

\paragraph{Rehearsal}
Rehearsal-based methods utilize a limited number of representative examples stored from old tasks or replay old examples using generative models while training new tasks.
Incremental Classifier Representation Learning (iCaRL)~\cite{rebuffi2017icarl} keeps a small number of samples per class to approximate the class centroid and makes predictions based on the nearest class mean classifiers.
On the other hand, pseudo-rehearsal techniques~\cite{ostapenko2019learning,shin2017continual} generate examples in the previously observed classes using generative adversarial networks (GANs)~\cite{goodfellow2014generative,liu2020generative,odena2017conditional}.
However, these methods need to train and store generative models, which incurs an extra burden for class incremental learning.

\paragraph{Adversarial attacks}
One can mislead deep neural networks trained on natural datasets by injecting minor perturbation into input data, which is called an adversarial attack~\cite{carlini2017cw,goodfellow2015fgsm,madry2018pgd,sey2016df,zhao2018nat}.
A well-known category in adversarial attacks relies on gradient-based optimization~\cite{carlini2017cw,goodfellow2015fgsm,madry2018pgd}. 
These methods deceive a network by adding a small amount of noise imperceptible by humans, in the direction of increasing the loss corresponding to the ground-truth label.
\begin{figure}[t]
  \centering
   \includegraphics[width=\linewidth]{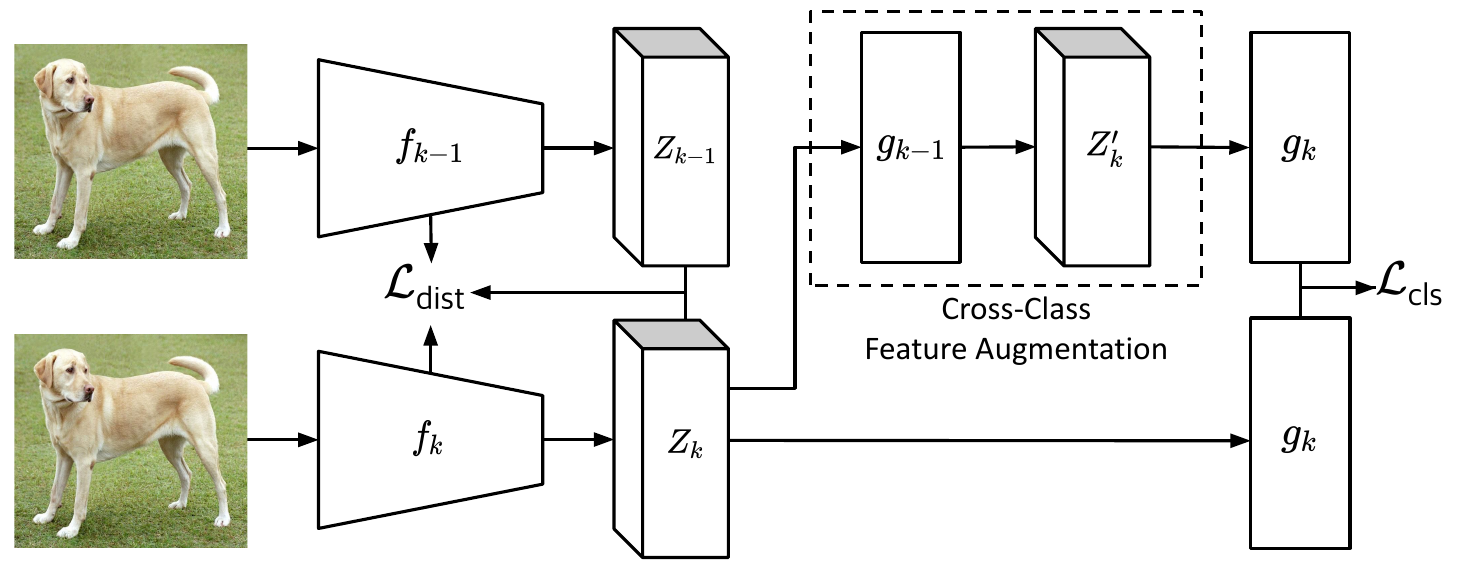}
   \caption{Overall class incremental learning framework with the proposed Cross-Class Feature Augmentation (CCFA). 
   Our model minimizes classification loss $\mathcal{L}_{\text{cls}}$ on training examples in a mini-batch sampled from the union of current task dataset and a small set of exemplars from the previous tasks while minimizing the distillation loss $\mathcal{L}_{\text{dist}}$.  
  To deal with the catastrophic forgetting issue induced by data imbalance between the previous and current tasks, we employ the CCFA to generate diverse features supporting the decision boundaries of the old classifier ${g}_{k-1}(\cdot)$. 
   }
   \vspace{-2mm}
   \label{fig:overall}
\end{figure}

\section{Proposed Approach}
\label{sec:method}
This section discusses the main idea and detailed algorithm of the proposed approach.

\subsection{Problem Setup}
\label{sec:problem_setup}

Class incremental learning trains a model in an online manner given a sequence of tasks, which is denoted by $T_{1:K} \equiv \{T_1, \cdots, T_k, \cdots, T_K \}$.
Each task $T_k$ is defined by a training dataset $\mathcal{D}_k$ composed of examples with labels $y \in \mathcal{Y}_k$, where $ (\mathcal{Y}_1 \cup \cdots \cup \mathcal{Y}_{k-1}) \cap \mathcal{Y}_k = \emptyset$. At the $k^\text{th}$ incremental stage, the model is trained on $\mathcal{D}'_{k}=\mathcal{D}_{k} \cup \mathcal{M}_{k-1}$, where $ \mathcal{M}_{k-1}$ is a small subset of all previously seen training datasets, which is called a memory buffer.
The performance of the trained model is evaluated on the test data sampled from a collection of all the encountered tasks without task boundaries. 

\subsection{Overall Framework}
\label{sec:method_overall}
We incorporate our feature augmentation technique into existing class incremental learning framework based on knowledge distillation~\cite{douillard2020podnet,rebuffi2017icarl, kang2022class}, and Figure~\ref{fig:overall} illustrates the concept of our approach.
At the $k^\text{th}$ incremental stage, we train the current model parametrized by ${\Theta}_k$, which consists of a feature extractor ${f}_k(\cdot)$ and a classifier ${g}_k(\cdot)$, using $\mathcal{D}'_{k}$.
The model optimizes ${\Theta}_k$ on a new task, $T_k$, initialized by the model parameters in the previous stage, ${\Theta}_{k-1}$, while preserving the information learned from old tasks, $T_{1:k-1}$, by using knowledge distillation.
To further enhance the generalization ability on the previously learned classes, we introduce Cross-Class Feature Augmentation (CCFA), which will be further discussed in the next subsection.
After each incremental stage, we sample exemplars from $\mathcal{D}_k$ by a herding strategy~\cite{rebuffi2017icarl} and augment the memory buffer from $\mathcal{M}_{k-1}$ to $\mathcal{M}_{k}$.
\begin{figure}[t]
  \centering
   \includegraphics[width=\linewidth]{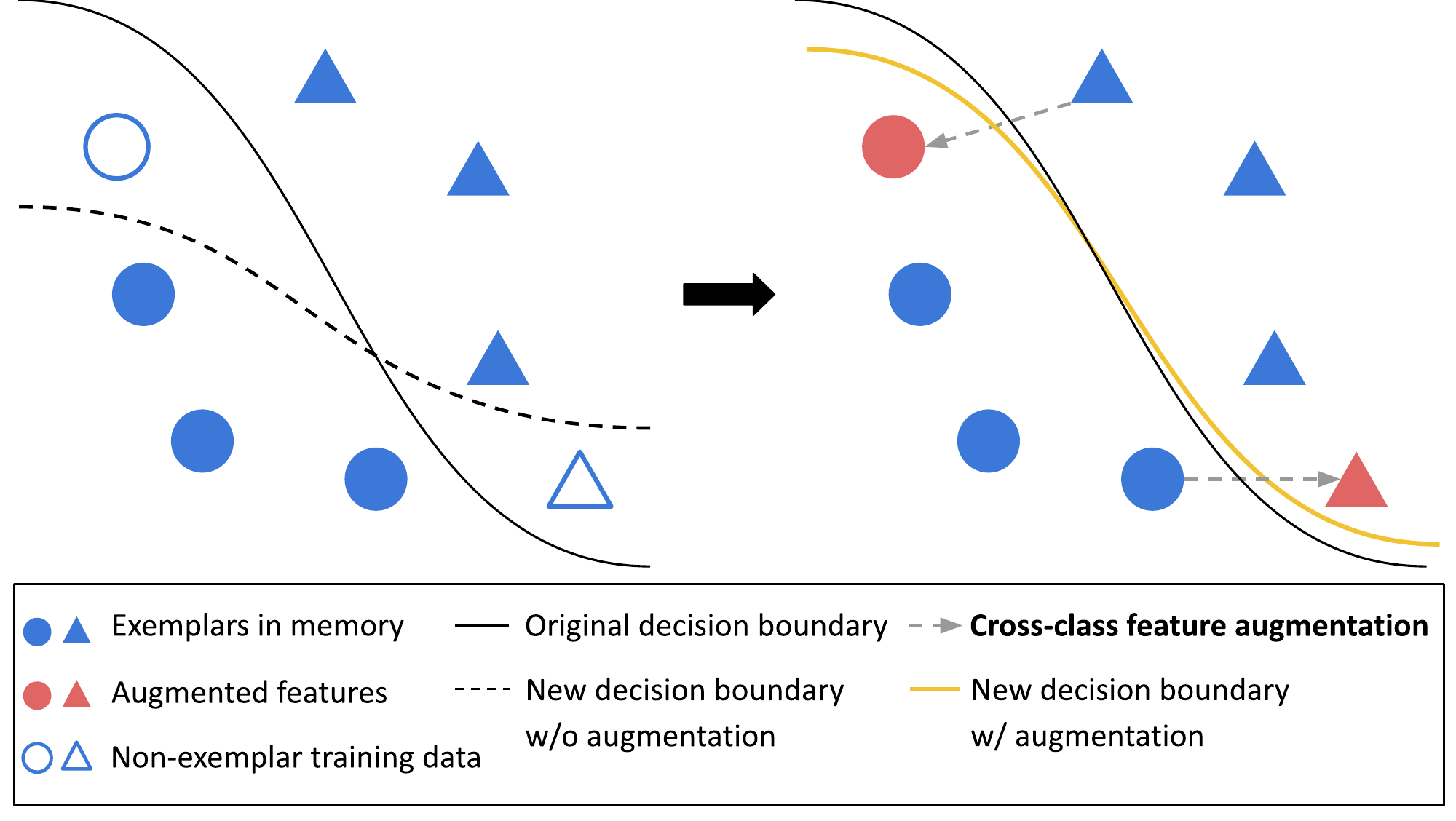}
   \caption{Illustration of Cross-Class Feature Augmentation (CCFA). 
CCFA perturbs a feature representation in a direction such that the perturbed feature crosses the decision boundary in the previous classifier to the target class, which is different from the original class, and complements the features for the target classes learned up to the previous stages.
      }
   \vspace{-2mm}
   \label{fig:ccfa}
\end{figure}
\subsection{Cross-Class Feature Augmentation (CCFA)}
\label{sec:method_ccfa}
Using a set of features extracted from the available training data, CCFA conducts feature augmentation to supplement training examples for the previous tasks. 
The feature augmentation is done in a similar way as adversarial attacks and does not require introducing and training a generator.
For CCFA, we utilize the classifier learned in the previous stage, which is readily accessible.
The main idea of CCFA is to perturb a feature representation in a direction such that the perturbed feature \textit{crosses} the decision boundary in the previous classifier $g_{k-1}(\cdot)$ to the target class.
Here, target classes are different from the original ones, as depicted in Figure~\ref{fig:ccfa}.


To be specific, for input images and the corresponding labels in a mini-batch $(X, Y) \sim \mathcal{D}'_{k}$, we first extract normalized feature vectors $Z_k = f_k(X)$. 
Then, we compute confidences of individual in-batch examples with respect to the class labels appearing in previous stages and define a collective confidence matrix $W \in \mathbb{R}^{b \times c_{\text{old}}}$ as follows:
\begin{equation}
W_{ij} = \begin{cases}
0, & \text{if } Y_{i} = j \\
W'_{ij}, & \text{if } Y_{i} \neq j,
\end{cases}
\end{equation}
where $b$ is the batch size, $c_{\text{old}} = |\mathcal{Y}_{1:k-1}|$ is the number of classes in previous stage, and $W'=g_k(Z_k)$ is the output of current classifier. 
Note that the entries for the ground-truth classes are set to zeros.

We formulate an optimization problem---selecting the target classes as evenly distributed as possible while maximizing the confidence scores---which is given by
\begin{equation}\label{eq:target}
\begin{split}
&\max_{S} \sum_{j=1}^{c_\text{old}} \sum_{i=1}^{b} W_{ij} S_{ij} - \sum_{j=1}^{c_\text{old}} \big \lvert \sum_{i=1}^{b} S_{ij}-u \big \rvert  \\
&\text{such that} \quad \ S \in \{0,1\}^{b \times c_{\text{old}}}, \ \sum_{j=1}^{c_\text{old}} S_{ij} =1 \  \ \forall \ i,
\end{split}
\end{equation}
where  $u = \frac{b}{c_\text{old}}$ is a constant. 
The solution of Eq.~\eqref{eq:target}, a matrix consisting of $b$ one-hot row vectors, provides a collection of target classes, $Y_\text{target}$.

Due to the computational complexity, we leverage a continuous relaxation of $S$ and reduce the number of active variables by selecting the top-$K$ classes with the highest confidence in each row of $W$.
Note that the entries in $S_i$ except the ones corresponding to the top-$K$ classes are simply set to zeros and target classes are to be sampled from the matrix $S$ whose rows act as sampling distributions.

Once the target classes are determined, we apply the Projected Gradient Descent (PGD)~\cite{madry2018pgd} algorithm iteratively using the following equation:
\begin{equation}\label{eq:attack process}
Z_{k}^{n+1} = Z_{k}^{n} - \alpha \sign (\nabla_{Z_k^n} \ell_\text{cls}(g_{k-1}(Z_k^n), Y_\text{target})), 
\end{equation}
where  $Z_k^0 = Z_k$, $\ell_\text{cls}(\cdot, \cdot)$ is the algorithm-specific classification loss function of each baseline, and $\alpha$ denotes the step size for the adversarial attack.
Note that we attack the normalized features just before the classification layer, unlike the original adversarial attack operating on the image space.

We obtain an adjusted features $Z_{k}'$ after performing the above process for $N$ steps. 
By passing $Z_{k}'$ through the old classifier $g_{k-1}(\cdot)$, we obtain pseudo-labels $Y_\text{pseudo}$ for $Z_{k}'$ as
\begin{equation}\label{eq:pseudo target selection}
\begin{split}
&Y_\text{pseudo} = \argmax g_{k-1}({Z'_k}) . \\
\end{split}
\end{equation}
%
We train the model with the pseudo-labeled augmented features in addition to the original ones, where the classification loss is given by
%
\begin{equation}\label{eq:classification}
\mathcal{L}_\text{cls} = \ell_\text{cls}([g_{k}(Z_{k}), g_{k}(Z_{k}')], [Y, Y_\text{pseudo}]),
\end{equation}
where $[\cdot, \cdot]$ denotes the concatenation operator along the batch dimension.

The unique aspect of CCFA is that it augments features across class boundaries.
Feature augmentation within the same classes could be another reasonable option to approximate the distribution of the old training dataset.
However, according to our experiments, this strategy suffers from the lack of diversity in the source of augmentation.
On the other hand, CCFA generates heterogeneous features from diverse sources in other classes, which handle sample deficiency in the previous tasks effectively.

\subsection{Training Objective}
\label{sec:method_objective}
The proposed approach is generic and can be integrated into various class incremental learning algorithms.
In this paper, we are interested in the methods based on knowledge distillation and the loss function at each stage is given by
\begin{equation}
\mathcal{L}_{\text{final}} = \mathcal{L}_\text{cls} + \lambda \cdot \mathcal{L}_\text{dist},
\label{eq:loss_definition}
\end{equation}
where $\mathcal{L}_\text{dist}$ is the distillation loss for each baseline algorithm, and $\lambda$ is the balancing weight. 
The augmented features given by CCFA only affect $\mathcal{L}_\text{cls}$, not $\mathcal{L}_\text{dist}$.

\section{Experiments}
\label{sec:experiments}

\subsection{Datasets and Evaluation Protocol}
\label{sub:datasets}
We evaluate the proposed framework on two datasets for class incremental learning, CIFAR-100~\cite{cifar09}, and ImageNet-100/1000~\cite{ILSVRC15}.
For fair comparison, we arrange the classes in three different orders for CIFAR-100 and in a single order for ImageNet-100 and ImageNet-1000 as provided by~\cite{douillard2020podnet}.
Following the previous works~\cite{douillard2020podnet,hou2019learning,liu2021adaptive}, we train the model using a half of the classes in the initial stage and split the remaining classes into groups of 50, 25, 10 and 5 for CIFAR-100, and 10 and 5 for ImageNet-100/1000.
At each incremental stage, we evaluate the model with the test examples in all the encountered classes. 
We report the average of the accuracies aggregated from all the incremental stages, which is also referred to as \textit{average incremental accuracy}~\cite{douillard2020podnet,hou2019learning,rebuffi2017icarl}.

\begin{table}[t!]
	\begin{center}
		\scalebox{0.8}{
			\setlength\tabcolsep{3pt}
			\hspace{-3mm}
			\begin{tabular}{lllll}
\multicolumn{1}{l}{Number of tasks}  & \multicolumn{1}{c}{50}      &  \multicolumn{1}{c}{25}   &  \multicolumn{1}{c}{10} &  \multicolumn{1}{c}{5}     \\ \midrule
				 iCaRL~\cite{rebuffi2017icarl}                   &    44.20             &  50.60 &  53.78      &   58.08                \\
				 BiC~\cite{wu2019large}                        &    47.09            &  48.96 &  53.21      &   56.86                 \\ 
				 Mnemonics~\cite{liu2020mnemonics}                        &     \hspace{0.34cm}-            &  60.96 &  62.28      &   63.34                 \\ 
				 GDumb{*}~\cite{prabhu2020gdumb}                        &    59.76            &  59.97 &  60.24      &   60.70                 \\ 
				 TPCIL~\cite{tao2020topology}                        &    \hspace{0.34cm}-          &  \hspace{0.34cm}- &  63.58      &   65.34                 \\ 
				 GeoDL{*}~\cite{simon2021learning}                        &    52.28            &  60.21 &  63.61      &   65.34                 \\ 
				  UCIR~\cite{hou2019learning}                   &    49.30     &  57.57     & 61.22            &  64.01            \\
				 \hline 
				 PODNet{}~\cite{douillard2020podnet}           &   57.98          &   60.72    &  63.19     &  64.83      \\ 
				  PODNet{*}~\cite{douillard2020podnet}           &   57.84            &   60.50      &  62.77     &  64.62       \\ 
				 PODNet{*}~\cite{douillard2020podnet}   + CCFA  &  \textbf{60.69}  &   \textbf{62.91}  &  \textbf{65.50}   & \textbf{67.24}    \\ \hline
				 AANet~\cite{liu2021adaptive}                   &  \hspace{0.34cm}-   &   62.31    &  64.31      &  66.31    \\ 
				 AANet{*}~\cite{liu2021adaptive}                   &  60.91 &   62.34     &  64.49     &  66.34      \\ 
				 AANet{*}~\cite{liu2021adaptive}  +  CCFA      & \textbf{62.20}    &\textbf{63.74}      &  \textbf{66.16}    &  \textbf{67.37} \\ \hline	 
				  AFC~\cite{kang2022class}                   &  62.18   &   64.06      &  64.29     &  65.82       \\ 
				  AFC{*}~\cite{kang2022class}                   &  61.74   &   63.78     &  64.63     &  66.02      \\ 
				 AFC{*}~\cite{kang2022class}  +  CCFA      & \textbf{63.11}    &   \textbf{64.59}    &  \textbf{65.61}     &  \textbf{66.47}  \\ 	 
				 \bottomrule				 
			\end{tabular} 
		}
	\end{center}
	\vspace{-2mm}
	\caption{Class incremental learning performance on CIFAR-100 for our model and the state-of-the-art frameworks.
	CCFA consistently improves the performance when plugged into the existing methods.
	Models with $*$ are our reproduced results. 
         Note that we run 3 experiments with 3 different orders for CIFAR-100 and report the average performance.	
	The bold-faced numbers indicate the best performance.}
	\label{tab:main_table_cifar}
	\vspace{-4mm}
\end{table}

\begin{table*}[t!]
	\begin{center}
		\scalebox{0.8}{
			\setlength\tabcolsep{6pt}
			\begin{tabular}{cclcccc}
  \multicolumn{1}{c}{Dataset} &  \multicolumn{1}{c}{Number of tasks} & \multicolumn{1}{c}{Memory per class $(m)$}& \multicolumn{1}{c}{$m$=1}   & \multicolumn{1}{c}{$m$=5} &  \multicolumn{1}{c}{$m$=\textbf{20}} \\ \midrule
	 	 \multirow{18}{*}{ImageNet-100} &  \multirow{9}{*}{5}	&  PODNet~\cite{douillard2020podnet}             & ---   &--- &  75.54 \\ 
		 
		        & &	PODNet{*}~\cite{douillard2020podnet}             & 50.18   &67.03 & 74.47   \\ 
			& &	 PODNet{*}~\cite{douillard2020podnet}   + CCFA      & \textbf{64.28}     & \textbf{72.42} & \textbf{75.78}  \\ \cline{3-6}
	                & &AANet~\cite{liu2021adaptive}             & ---    &--- &  76.96 \\ 
			& &	 AANet{*}~\cite{liu2021adaptive}            & 71.25   &75.22 & 78.05   \\ 
			& &	 AANet{*}~\cite{liu2021adaptive}  + CCFA      & \textbf{74.48}     & \textbf{76.75} & \textbf{78.14}  \\ \cline{3-6}
			&   & AFC~\cite{kang2022class}             & --   &-- & 76.87  \\ 
			& &	 AFC{*}~\cite{kang2022class}             & 56.53   &72.75 & \textbf{76.91}  \\ 
			& &	 AFC{*}~\cite{kang2022class}   + CCFA      & \textbf{64.45}    & \textbf{74.35} & 76.75  \\ \cline{2-6}
			&  \multirow{9}{*}{10}	&	 PODNet~\cite{douillard2020podnet}             & ---   & --- & 74.33   \\ 

			& &PODNet{*}~\cite{douillard2020podnet}        & 37.64     &63.21 & 72.37 \\ 
			& & PODNet{*}~\cite{douillard2020podnet}   + CCFA    & \textbf{49.80}    &\textbf{65.62}  & \textbf{73.00}  \\  \cline{3-6}
	 	 	& & AANet~\cite{liu2021adaptive}             & ---     & --- & 74.33   \\ 
			& & AANet{*}~\cite{liu2021adaptive}        & 60.74     &71.21 & 75.90 \\ 
			& & AANet{*}~\cite{liu2021adaptive}  + CCFA    & \textbf{66.62}     &\textbf{73.51}& \textbf{76.71} \\  \cline{3-6}
			 & &AFC~\cite{kang2022class}             & --   & -- & 75.75\\ 
			& &	 AFC{*}~\cite{kang2022class}             & 51.37   &70.69 & 75.10   \\ 
			 &&	 AFC{*}~\cite{kang2022class}   + CCFA      & \textbf{61.85}     & \textbf{71.35} & \textbf{75.39} \\
				 \midrule
			\multirow{12}{*}{ImageNet-1000} & \multirow{6}{*}{5}	&	 PODNet~\cite{douillard2020podnet}             & ---   &  ---  &  66.95 \\ 
			& & PODNet{*}~\cite{douillard2020podnet}             & 51.20  & 66.08 & 69.70   \\ 
			& &	 PODNet{*}~\cite{douillard2020podnet}   + CCFA      & \textbf{59.78}  & \textbf{69.03}  & \textbf{69.97} \\  \cline{3-6} 
			&&AFC~\cite{kang2022class}             & -- & -- & 67.02   \\ 
			&&	 AFC{*}~\cite{kang2022class}             & 57.37  & 66.47 & 67.93   \\ 
			 &&	 AFC{*}~\cite{kang2022class}   + CCFA      & \textbf{65.11}  & \textbf{68.78}  & \textbf{69.55} \\ \cline{2-6}
			&  \multirow{6}{*}{10}	&	 PODNet~\cite{douillard2020podnet}             & ---    & --- & 64.13   \\ 
			&&PODNet{*}~\cite{douillard2020podnet}        & 36.66    &  59.34  & 67.11 \\ 
			& & PODNet{*}~\cite{douillard2020podnet}   + CCFA    & \textbf{48.41}    &\textbf{63.74}  &\textbf{67.82} \\ 	 \cline{3-6} 
			 &&AFC~\cite{kang2022class}        & --&-- & 67.02 \\ 
			& & AFC{*}~\cite{kang2022class}        & 52.36   & 63.62  & 65.96 \\ 
			 && AFC{*}~\cite{kang2022class}   + CCFA    & \textbf{63.42}    &\textbf{67.07}  &\textbf{67.88} \\ 	
			\bottomrule
			\end{tabular} 
		}
	\end{center}
	\vspace{-2mm}
	\caption{Class incremental learning performance on ImageNet-100 and ImageNet-1000 for our model and the baseline algorithms with varying memory sizes. 
	CCFA consistently improves the performance when plugged into the baselines, especially with an extremely limited memory budget.
	Models with an asterisk ($*$) are our reproduced results. 
	The bold-faced numbers indicate the best performance.}
	\label{tab:main_table_imagenet100}
	\vspace{-2mm}
\end{table*}

\subsection{Implementation Details}
\label{sub:imp_details} 
We follow the implementation settings of the existing methods~\cite{douillard2020podnet,liu2021adaptive, kang2022class}. 
We adopt ResNet-32 for CIFAR-100 and ResNet-18 for ImageNet as the backbone network architectures. 
The hyperparameters including learning rates, batch sizes, training epochs, distillation loss weight ($\lambda$) and herding strategies are identical to the baseline algorithms~\cite{douillard2020podnet,liu2021adaptive, kang2022class}.  
The size of the memory buffer is set to 20 per class as a default for all experiments unless specified otherwise.

For the feature augmentation, we set the number of iterations for adversarial attack to 10 for all experiments.
Empirically, we find that $K=1$ is sufficient for top-K in target selection which is equivalent to selecting classes with highest confidence as $Y_{\text{target}}$ without solving LP optimization problem. 
In the CIFAR-100 experiments, we randomly sample the attack step size $\alpha$ from a uniform distribution $\mathcal{U}(\frac{2}{255},\frac{5}{255})$ and generate 640 features, which is 5 times the batch size, using 5 randomly sampled values of $\alpha$  for each real example.
For ImageNet, the attack step size $\alpha$ is randomly sampled from the uniform distribution $\mathcal{U}(\frac{2}{2040},\frac{5}{2040})$ and 128 features, which is equal to the batch size, are generated using a random sample of $\alpha$.
We select the step size $\alpha$ in a way that the same amount of feature updates occurs in a single iteration for backbone networks with different feature dimensions.
For the number of augmented features, we consider the stability-plasticity trade-off.

\subsection{Results on CIFAR-100}
\label{sub:results_cifar}
We compare the proposed approach, referred to as Cross-Class Feature Augmentation (CCFA), with existing state-of-the-art class incremental learning methods.
We incorporate CCFA into four baseline models including PODNet~\cite{douillard2020podnet}\footnote{https://github.com/arthurdouillard/incremental\_learning.pytorch}, AANet~\cite{liu2021adaptive}\footnote{https://github.com/yaoyao-liu/class-incremental-learning}, and AFC~\cite{kang2022class}\footnote{https://github.com/kminsoo/AFC}. 
Note that we run three experiments with three different orders for CIFAR-100 and report the average performance.
Table~\ref{tab:main_table_cifar} demonstrates that the proposed method consistently improves accuracy on three baseline models in various scenarios, and achieves the state-of-the-art performance.

\subsection{Results on ImageNet} 
\label{sub:results_imagenet}
We test the performance of CCFA on ImageNet-100/1000 with varying size of the memory buffer.
Table~\ref{tab:main_table_imagenet100} presents the results of our method compared to the baselines, PODNet~\cite{douillard2020podnet}, AANet~\cite{liu2021adaptive}, and AFC~\cite{kang2022class} by varying the number of tasks and the allocated memory per task.
CCFA boosts the performances consistently, especially when the size of the memory buffer is extremely small.
These results show that the features augmented by CCFA play a crucial role in complementing the lack of training examples in old tasks and alleviating the collapse of the decision boundaries.
Moreover, CCFA is helpful for regularizing the model by augmenting diverse samples and reducing overfitting issue commonly observed when the memory size is small.

\subsection{Ablation Study and Analysis}
\label{sub:ablation}
We perform several ablation studies on CIFAR-100 to analyze the effectiveness of CCFA.
For all the ablation studies, we utilize PODNet as the baseline.
The number of incremental stages is set to 50 unless specified otherwise.

\begin{table}[t!]
\begin{center}
	\scalebox{0.8}{
	\setlength\tabcolsep{5pt}
\begin{tabular}{llc}
Ablation types &  Variations	& Acc (\%)   \\ 
\midrule
\multirow{2}{*}{(a) Initialization}  & Random noise       &    59.90        \\
\multicolumn{1}{c}{}               &   $Z_k$  (ours) 		&    \textbf{60.69}   \\ \midrule
\multirow{3}{*}{(b) Target class}  & Random      &   59.91    \\
\multicolumn{1}{c}{}               &   Farthest     & 57.38  \\
 \multicolumn{1}{c}{}               &   Nearest  (ours)   & \textbf{60.69}     \\ \midrule
 \multirow{4}{*}{\shortstack[l]{(c) Top-$K$ \\ \quad(5 stages)}}  & $K=10$     &  \textbf{67.56}    \\
 \multicolumn{1}{c}{}               &   $K=5$     & 67.39    \\
 \multicolumn{1}{c}{}               &   $K=3$   & 67.38    \\                               
\multicolumn{1}{c}{}               &   $K=1$ (ours)     & 67.24  \\ \midrule
\multirow{6}{*}{\shortstack[l]{(d) Augmentation \\ \hspace{4mm} methods}}  & PODNet~\cite{douillard2020podnet}      &   57.98    \\
\multicolumn{1}{c}{}                   &   CutMix	~\cite{yun2019cutmix}  &    58.92          \\
\multicolumn{1}{c}{}                  &    MixUp~\cite{zhang2017mixup}    &  55.96    \\ 
\multicolumn{1}{c}{}                  &    Manifold-MixUp~\cite{verma2019manifold}    &  55.14    \\ 
\multicolumn{1}{c}{}                  &    CCFA  		&    \textbf{60.69}   \\ \midrule
\multirow{2}{*}{\shortstack[l]{(e) Exemplar-free \\ \quad(10 stages)}}                  &    IL2A~\cite{zhu2021class}  		&   58.42   \\
\multicolumn{1}{c}{}                  &    IL2A~\cite{zhu2021class} + CCFA  		&    \textbf{60.78}   \\
\bottomrule
\end{tabular}
}
\end{center}
\vspace{-2mm}
\caption{Ablation study results on the variations of CCFA.
	The bold-faced numbers indicate the best performance.}
\label{tab:multi_abl}
\vspace{-2mm}
\end{table}
\paragraph{\bf Initialization for augmentation}
\label{sub:ablation_augment}
Table~\ref{tab:multi_abl}(a) presents the comparison between random noise and extracted feature, $z_k$, as an initialization point for perturbation.
To implement the random initialization, we replace $Z_k$ in Equation~\eqref{eq:attack process} by a sample from a Gaussian distribution, $\mathcal{N}(\textbf{0}, \mathbf{I})$.
According to our experiment, the augmented feature from $Z_k$ outperforms the random noise while the random initilization is still helpful for boosting performance.

\paragraph{\bf Target class selection strategy}
\label{sub:ablation_target}
We evaluate the proposed cross-class feature augmentation with different target selection criterion.
Table~\ref{tab:multi_abl}(b) demonstrates that CCFA boosts the baseline even with the random target class selection strategy since random target classes provide diverse augmentation directions.
However, the gradients toward low-confidence classes may be unstable, reducing the benefit of CCFA. 
Performance drop with the farthest target class supports this assumption.
In addition, we show the results with varying $K$ for top-$K$ used in target selection process.
Table~\ref{tab:multi_abl}(c) illustrates that setting $K=1$, \ie no optimization process added, shows comparable result with the results on larger $K$'s.
While increased $K$ shows improved results, introducing LP in batch training incurs excessive extra training costs.
On a single NVIDIA Titan Xp GPU, training with $K=1$ runs at 5.15 iteration/sec, which is 10 times faster than the case with $K=3$, when the ResNet-32 backbone is employed with batch size of 128.

\begin{table}[t!]
\centering
\setlength\tabcolsep{2pt}
\scalebox{0.8}{
\begin{tabular}{lcccc}
Number of tasks & 50 & 25 & 10 & 5 \\
\toprule
PODNet + Mixup~\cite{zhang2017mixup} & 55.96 & 59.22 & 63.97 & 66.26 \\
PODNet + Mixup~\cite{zhang2017mixup} + CCFA & \textbf{58.47} & \textbf{64.05} & \textbf{67.77} & \textbf{69.19} \\ \midrule
PODNet + CutMix~\cite{yun2019cutmix} & 58.92 & 62.69 & 66.18 & 68.30 \\
PODNet + CutMix~\cite{yun2019cutmix} + CCFA & \textbf{61.41} & \textbf{64.88} & \textbf{67.81} & \textbf{69.28} \\
\bottomrule
\end{tabular}
}
\vspace{-2mm}
\caption{Compatibility of CCFA with existing data augmentation techniques. The bold-faced numbers indicate the best performance}
\label{tab:mix}
\vspace{-2mm}
\end{table}

\paragraph{\bf Comparison with other data augmentation techniques} 
Data augmentation is a widely used method to increase the number of training examples and learn a robust model.
We show the superiority of CCFA to the standard data augmentation methods.
We employ CutMix~\cite{yun2019cutmix}, Mixup~\cite{zhang2017mixup} and Manifold-MixUp~\cite{verma2019manifold}.
Table~\ref{tab:multi_abl}(d) exhibits that CCFA clearly outperforms the existing augmentation methods for class incremental learning.

\paragraph{\bf Results on exemplar-free setting}
Table~\ref{tab:multi_abl}(e) presents the results of IL2A~\cite{zhu2021class} on CIFAR-100 with 10 incremental stages. IL2A~\cite{zhu2021class} is an exemplar-free method that employ two different augmentation strategies, class and semantic augmentations. 
Table~\ref{tab:multi_abl}(e) shows that CCFA boosts the performance of the baseline in exemplar-free settings. 
Note that semantic augmentation of IL2A~\cite{zhu2021class} samples features from a class-wise Gaussian distribution as in PASS~\cite{zhu2021prototype}.

\begin{table}[!t]
	\begin{center}
		\scalebox{0.8}{
			\setlength\tabcolsep{6pt}
			\hspace{-2mm}
			\begin{tabular}{llllllllll}
                   \multicolumn{1}{c}{\# of augmented features} & \multicolumn{1}{c}{1} &  \multicolumn{1}{c}{3} &  \multicolumn{1}{c}{5} &  \multicolumn{1}{c}{7} &\multicolumn{1}{c}{9} \\ \midrule
			Forgetting $\downarrow$ &	       28.13 & 26.19 & 24.97& 25.00  & 23.85  \\ 
			Average new accuracy $\uparrow$ &      76.58     & 72.11 & 68.94 &66.63 &65.15   \\ \midrule
			Overall accuracy &      59.14     & 60.17 & \textbf{60.69} &60.51 &60.51 \\
			 \bottomrule				 
			\end{tabular} 
		}
	\end{center}
	\vspace{-2mm}
	\caption{Forgetting vs adaptivity by varying the number of augmented features ($Z'_k$).
	The bold-faced numbers indicate the best performance.
	}
	\label{tab:batch}
\vspace{-2mm}
\end{table}

\paragraph{\bf Compatibility with the existing augmentation methods}
We investigate the compatibility of CCFA with the existing augmentation methods, Mixup~\cite{zhang2017mixup} and CutMix~\cite{yun2019cutmix}, to show broad applicability of CCFA.
Table~\ref{tab:mix} shows that CCFA benefits from the existing input-level augmentation techniques in every class incremental learning scenario.

\paragraph{\bf Number of augmented features}
\label{sub:ablation_number}
We conduct experiments by varying the number of augmented features. 
Since the quantity of the features affects the relative data ratio between the old and new tasks, we focus on the balance between learning new classes and forgetting old classes.
We measure adaptivity of the model to new classes using the average new accuracy.
For robustness of the model to old classes, we measure the forgetting metric~\cite{lee2019overcoming}.
Table~\ref{tab:batch} illustrates the performance of PODNet with CCFA in terms of the two metrics and the overall accuracy by varying the number of augmented features.
We observe that increasing the number of augmented features tends to reduce both forgetting and average new accuracy since they impose more weight on old classes.

\begin{table}[!t]
\begin{center}
  \scalebox{0.8}{
  	\setlength\tabcolsep{5pt}
	\hspace{-2mm}
 \begin{tabular}{lccccc}
Memory per class ($m$) & $m$=5     & $m$=10    & $m$=20    & $m$=50    \\
 \midrule
iCaRL \cite{rebuffi2017icarl}      & 16.44 & 28.57 & 44.20 & 48.29  \\
BiC~\cite{wu2019large}       & 20.84  & 21.97  & 47.09  & 55.01  \\
UCIR~\cite{hou2019learning} & 22.17 & 42.70 & 49.30 & 57.02 \\
PODNet~\cite{douillard2020podnet}  & 35.59 & 48.54 & 57.98 & 63.69  \\
\midrule
PODNet + CCFA (ours) & \textbf{39.70}  &  \textbf{52.25} & \textbf{60.69} &   \textbf{65.99}\\
\bottomrule
\end{tabular}
}
\end{center}
\vspace{-2mm}
\caption{Comparative results by varying the memory budget for each class on CIFAR-100 with 50 stages.
The results demonstrate the robustness of our algorithm with respect to memory budgets. 
}
\vspace{-2mm}
\label{tab:ablation_memory}
\end{table}

\paragraph{\bf Memory size}
\label{sub:memory}
We claim that the proposed method effectively compensate for the lack of training data at the feature space level. 
To validate this hypothesis, we analyze how the size of the memory budget affect the performance of our algorithm in Table~\ref{tab:ablation_memory}.
The results show that the accuracy gains obtained from CCFA indeed increase by decreasing the memory size, which is consistent with our assumption.

\begin{table}[!t]
\begin{center}
\scalebox{0.8}{
\setlength\tabcolsep{5pt}
\hspace{-2mm}
\begin{tabular}{lcccc}
\multicolumn{1}{l}{Initial task size} & 20 & 30 & 40 & 50\\ \midrule
\multicolumn{1}{l}{Number of stages} & 80  & 70  & 60  & 50  \\
 \midrule
iCaRL \cite{rebuffi2017icarl}     & 41.28 & 43.38 & 44.35 & 44.20\\
BiC \cite{wu2019large}       & 40.95  & 42.27 & 45.18 & 47.09\\
UCIR \cite{hou2019learning} & 41.69 & 47.85 & 47.51 & 49.30\\
PODNet~\cite{douillard2020podnet}  & 47.68 & 52.88 & 55.42 & 57.98\\
\midrule
PODNet + CCFA (ours) &  \textbf{50.32} &\textbf{55.51}  & \textbf{57.59} & \textbf{60.69} \\
\bottomrule
\end{tabular}
}
\end{center}
\vspace{-2mm}
\caption{Performance comparison between the proposed CCFA and the state-of-the-art frameworks on CIFAR-100 by varying the number of classes in the initial task while each of the remaining tasks only contains a single class. The bold-faced numbers represent the best performance.}
\label{tab:varying_initial_task}
\end{table}

\begin{table}[t!]
\begin{center}
\scalebox{0.8}{
\setlength\tabcolsep{7pt}
\hspace{-2mm}
\begin{tabular}{lccccc}
Memory per class ($m$) & $m$=1     & $m$=5 & $m$=10   & $m$=20      \\
 \midrule
PODNet~\cite{douillard2020podnet}  & 35.18 & 55.30  & 59.16 &  63.19  \\
PODNet + CCFA towards GT & 30.07  &57.13& 62.09 &  64.84 \\
\midrule
PODNet + CCFA (ours) & \textbf{38.06}  &\textbf{58.78} & \textbf{62.85} &  \textbf{65.50} \\
\bottomrule
\end{tabular}
}
\end{center}
\vspace{-2mm}
\caption{Analysis regarding the direction of the augmentation process on CIFAR-100 with 10 stages.
The results demonstrate effectiveness of setting the target class in a cross-class manner with various memory budgets. 
The bold-faced numbers represent the best performance.
}
\vspace{-2mm}
\label{tab:ccfa_gt}
\end{table}

\begin{figure}[t]
  \centering
  \includegraphics[width=\linewidth]{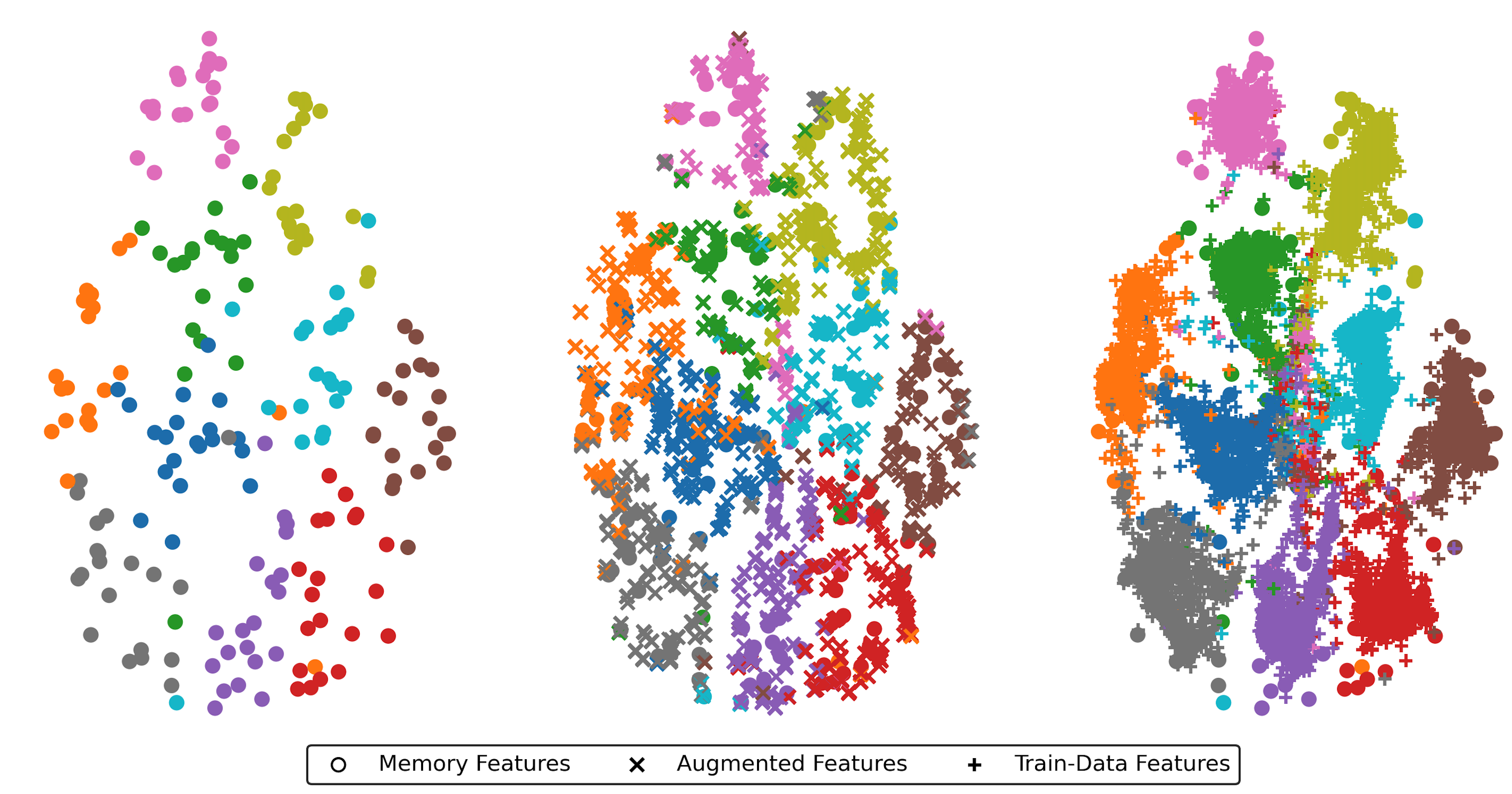}
  \vspace{-2mm}
   \caption{t-SNE of the features from the memory buffer of the random 10 classes after training initial stage (Memory Features), features generated by CCFA (Augmented Features) and features from randomly selected 1000 images from training dataset (Train-Data Features). By allowing CCFA, each class in the old tasks populates samples in the feature space, which alleviates the collapse of the decision boundaries caused by sample deficiency for the old tasks.
    }
   \label{fig:tsne}
\end{figure}

\paragraph{\bf Initial task size}
\label{sub:ablation_initial_task}
%
We evaluate CCFA in the settings, where the initial task is small and the initially learned features are not sufficiently robust.
Table~\ref{tab:varying_initial_task} illustrates the results with respect to diverse initial task sizes; our approach outperforms the baselines regardless of initial task sizes.
\vspace{-1mm}

\paragraph{\bf Cross-class augmentation strategy}
As discussed earlier, one reasonable augmentation strategy is to adopt the gradient direction that preserves the original class label and reduces the classification loss.
We compare CCFA with CCFA towards the ground-truth labels (CCFA towards GT) to demonstrate the efficiency of the proposed algorithm with various memory budgets in Table~\ref{tab:ccfa_gt}. 
Note that CCFA outperforms CCFA towards GT in all settings and their performance gap increases as the memory size gets smaller.
This result implies that the cross-class augmentation strategy in CCFA yields more diversity in the synthesized representations and alleviates the lack of augmentation sources in a limited memory environment.

\paragraph{\bf Computational complexity}
\label{sub:cost}
CCFA incurs a small amount of additional computation because the gradients are computed with respect to the classification layer only.
On a single NVIDIA RTX GPU, PODNet with CCFA requires 1.53 seconds per iteration while PODNet requires 1.50s with batch size 128 under the ResNet-18 backbone. 

\begin{figure}[t]
    \centering
    
    \begin{subfigure}{0.24\textwidth}
        \centering
        \includegraphics[width=\linewidth]{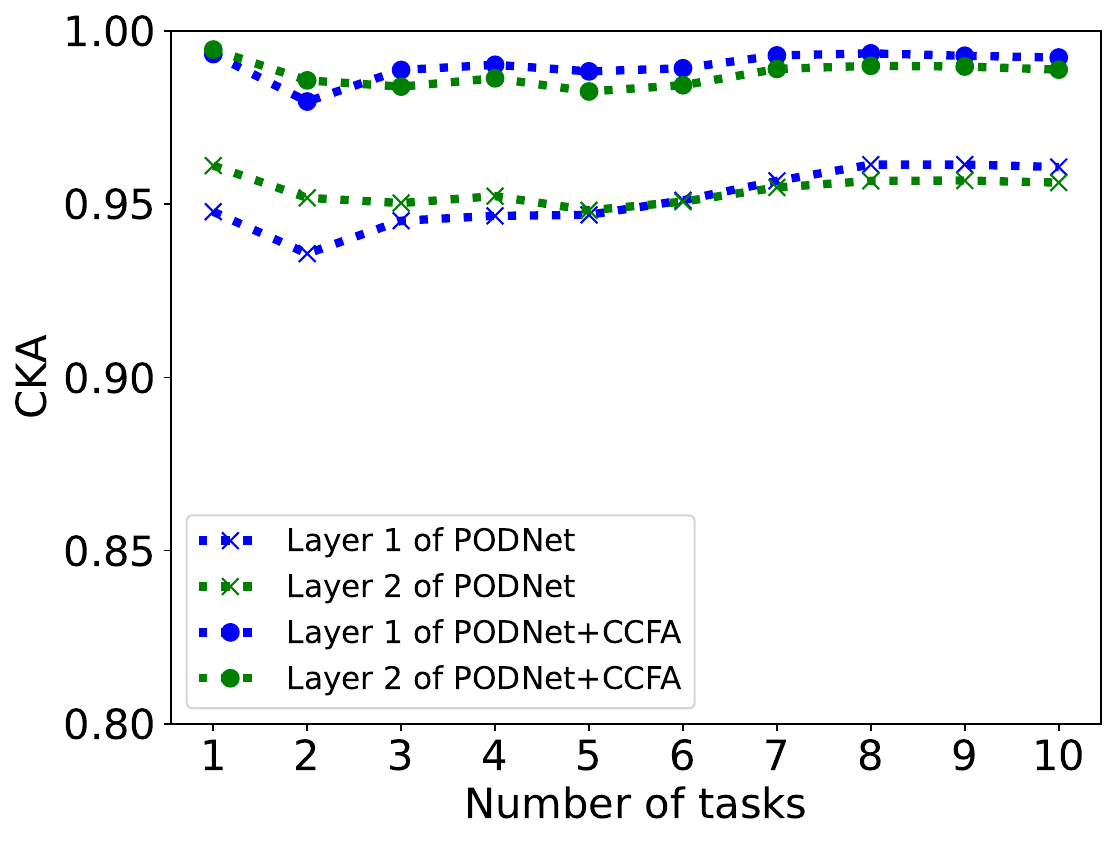}
        \vspace{-4mm}
        \caption{CKA after $1^\text{st}$/$2^\text{nd}$ layer}
        \label{subfig:1}
    \end{subfigure}%
    \begin{subfigure}{0.24\textwidth}
        \centering
        \includegraphics[width=\linewidth]{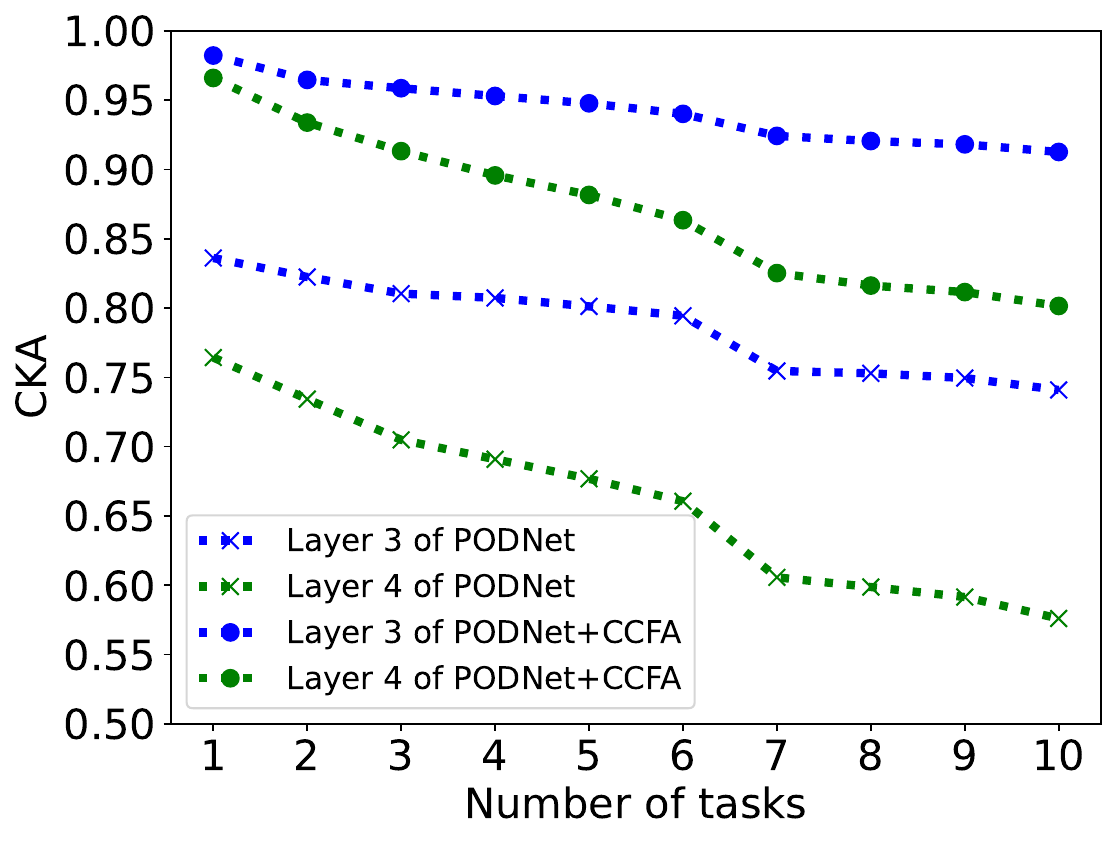}
        \vspace{-4mm}
        \caption{CKA after $3^\text{rd}$/$4^\text{th}$ layer}
        \label{subfig:2}
    \end{subfigure}
    \vspace{-2mm}
    \caption{CKA between the features from the feature extractors of the first and each incremental stage for PODNet~\cite{douillard2020podnet} and PODNet + CCFA after each residual layer.}
    \vspace{-4mm}
    \label{fig:cka}
\end{figure}

\subsection{Visualization of CCFA}
\label{sub:analysis}
To support our argument that augmented features given by CCFA resolve the sample deficiency problem of class incremental learning, we visualize the features from the memory buffer, augmented features, and features from training data that are inaccessible in subsequent stages.
Figure~\ref{fig:tsne} demonstrates that CCFA effectively approximates the structure of feature space given by full training data, which is consistent with our quantitative results.. 

\subsection{Effects of CCFA on the Feature Extractor}
\label{sub:feature}
To understand the impact of CCFA on the feature extractor, we measure the Centered Kernel Alignment (CKA)~\cite{cortes2012algorithms, kornblith2019similarity, kim2023stability} between the intermediate features of the feature extractor after the first incremental stage and each sequential incremental stage, with and without CCFA. 
We utilize the test set of the first stage to measure the drift of the feature representations learned in the first incremental stage.
Figure~\ref{fig:cka} illustrates the change of CKA values for each residual layer as the number of stages increases.
The results clearly demonstrate that CCFA alleviates forgetting at every intermediate feature level even though it only operates at the classifier level.

\section{Conclusion}\label{sec:conclusion}	
We presented a novel feature augmentation technique for class incremental learning based on adversarial attacks.
The proposed method augments the features of the target class by utilizing the samples from other classes, adversarially attacking the previously learned classifier.
The generated features play a role in complementing the data for the previous tasks, which become powerful supporting samples for the decision boundaries.
This idea is generally applicable to class incremental learning frameworks based on knowledge distillation without any modification on the architecture.
Our extensive experimental results demonstrate that the proposed algorithm consistently improves performance on multiple datasets when applied to existing class incremental learning frameworks, especially in an environment with extremely limited memory constraints.

\section*{Acknowledgements}
This work was partly supported by Samsung Advanced Institute of Technology, and the National Research Foundation of Korea grant [No. 2022R1A2C3012210, Knowledge Composition via Task-Distributed Federated Learning] and the IITP grants [No.2022-0-00959, (Part 2) Few-Shot Learning of Causal Inference in Vision and Language for Decision Making; 2021-0-02068, Artificial Intelligence Graduate School Program (Seoul National University); 2021-0-01343, Artificial Intelligence Innovation Hub] funded by the Korean government (MSIT).

\bibliography{aaai24}

\end{document}


\maketitle

\begin{figure}[t!]
 \hspace*{-3mm}
     \centering
     \begin{subfigure}[b]{0.45\linewidth}
         \centering
          \hspace*{-3mm}
           \includegraphics[width=1.1\linewidth]{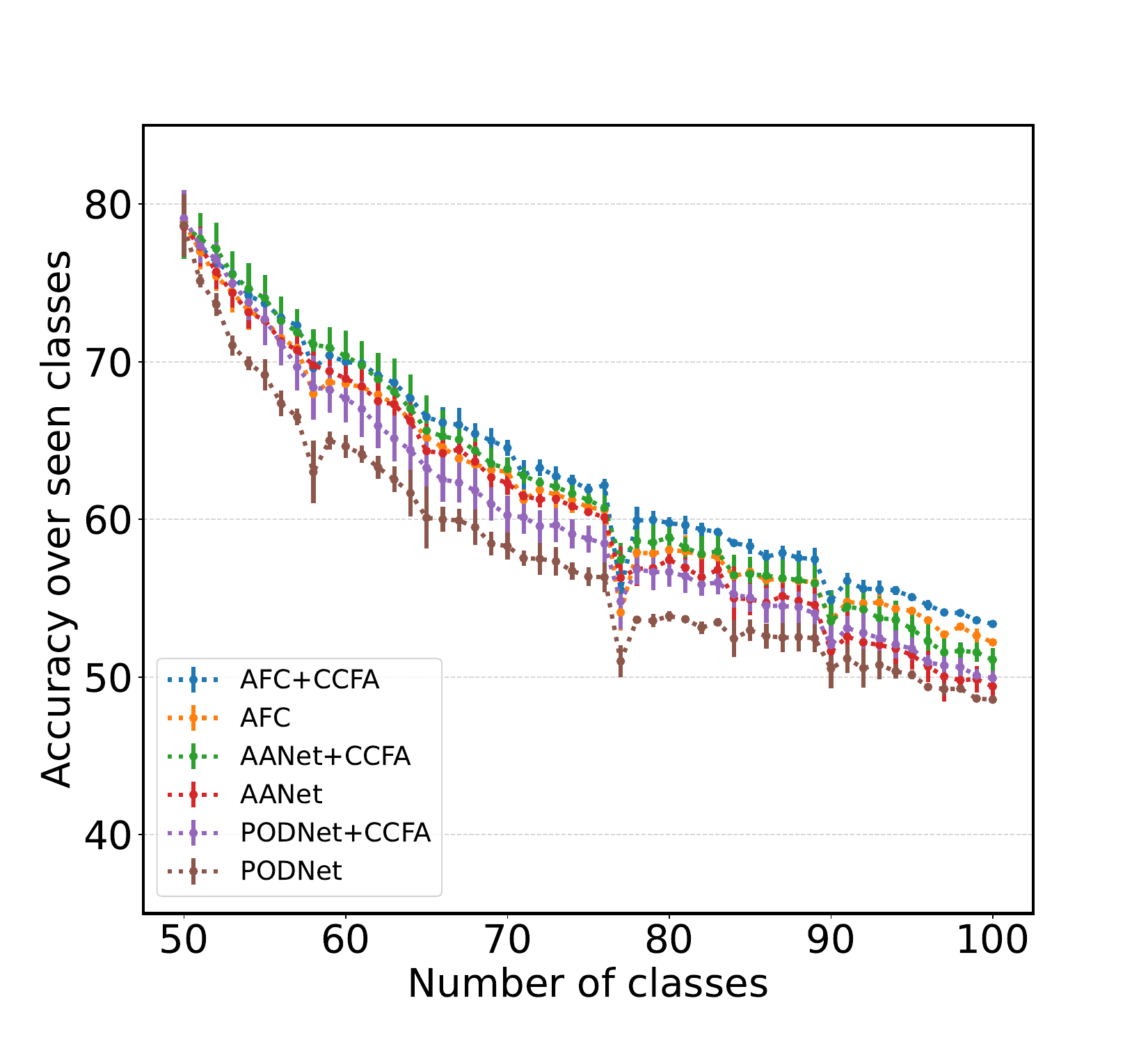}
           \vspace{-0.5cm}
          \caption{1$\times$50 stages}
         \label{fig:cifar50step}
     \end{subfigure}
     \vspace{0.2cm}
     \hspace{0.3cm}
     \begin{subfigure}[b]{0.45\linewidth}
         \centering
          \hspace*{-3mm}
         \includegraphics[width=1.1\linewidth]{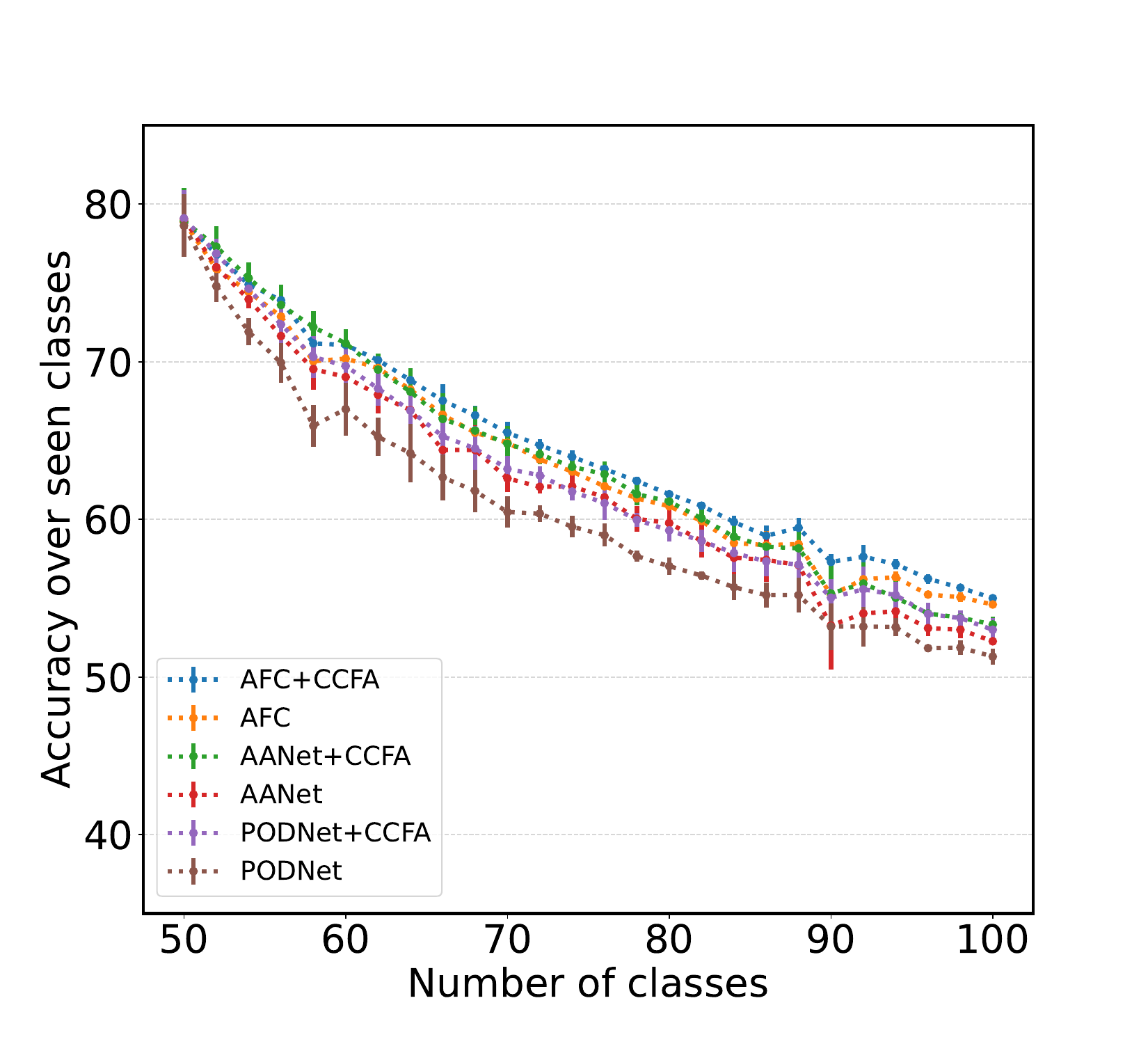}
          \vspace{-0.5cm}
         \caption{2$\times$25 stages}         
         \label{fig:cifar25step}
     \end{subfigure}
  
     \begin{subfigure}[b]{0.45\linewidth}
         \centering
          \hspace*{-3mm}
         \includegraphics[width=1.1\linewidth]{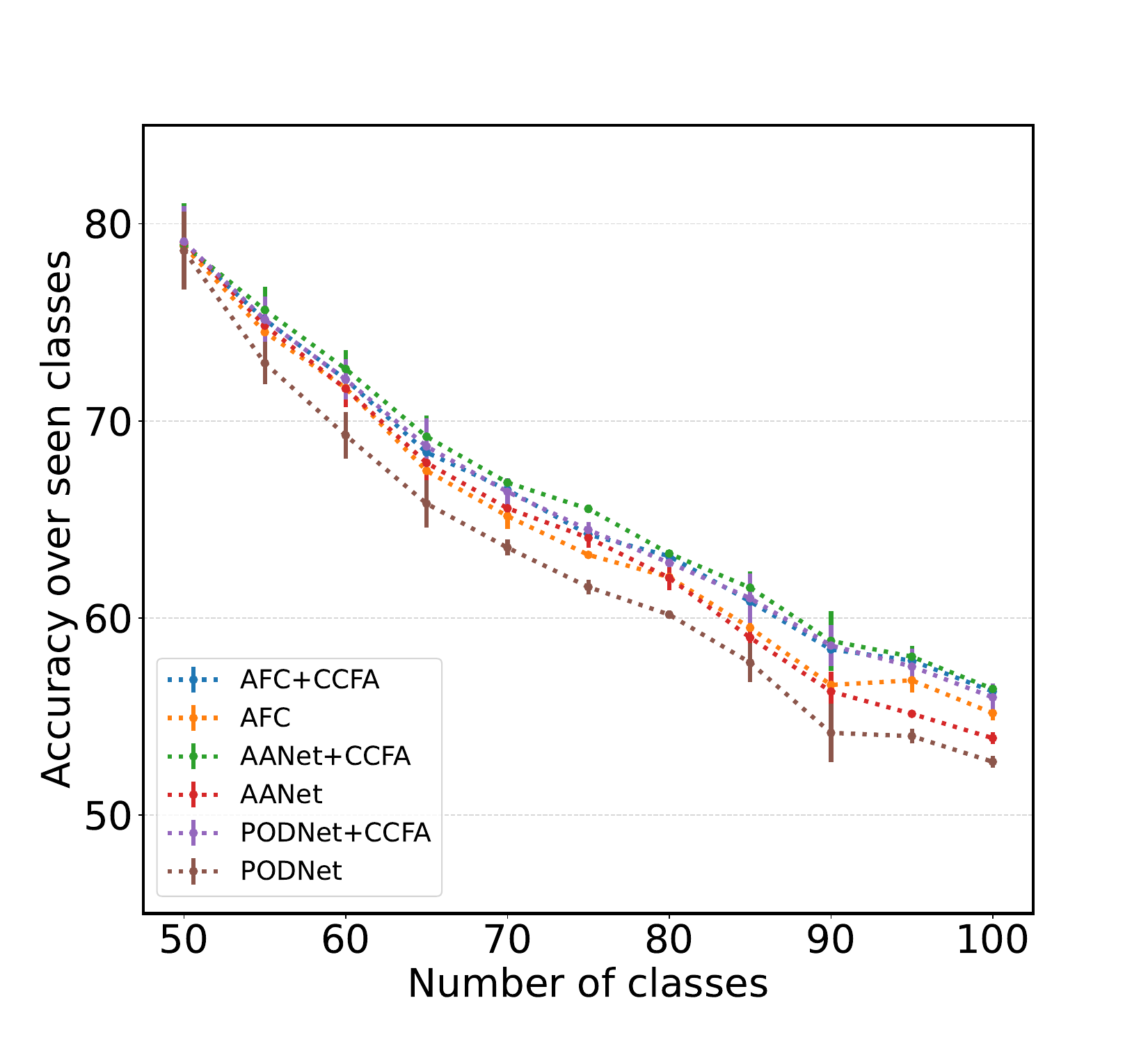}
          \vspace{-0.5cm}
         \caption{5$\times$10 stages}         
         \label{fig:cifar10step}
     \end{subfigure}
     \hspace{0.3cm}
     \begin{subfigure}[b]{0.45\linewidth}
         \centering
          \hspace*{-3mm}
          \includegraphics[width=1.1\linewidth]{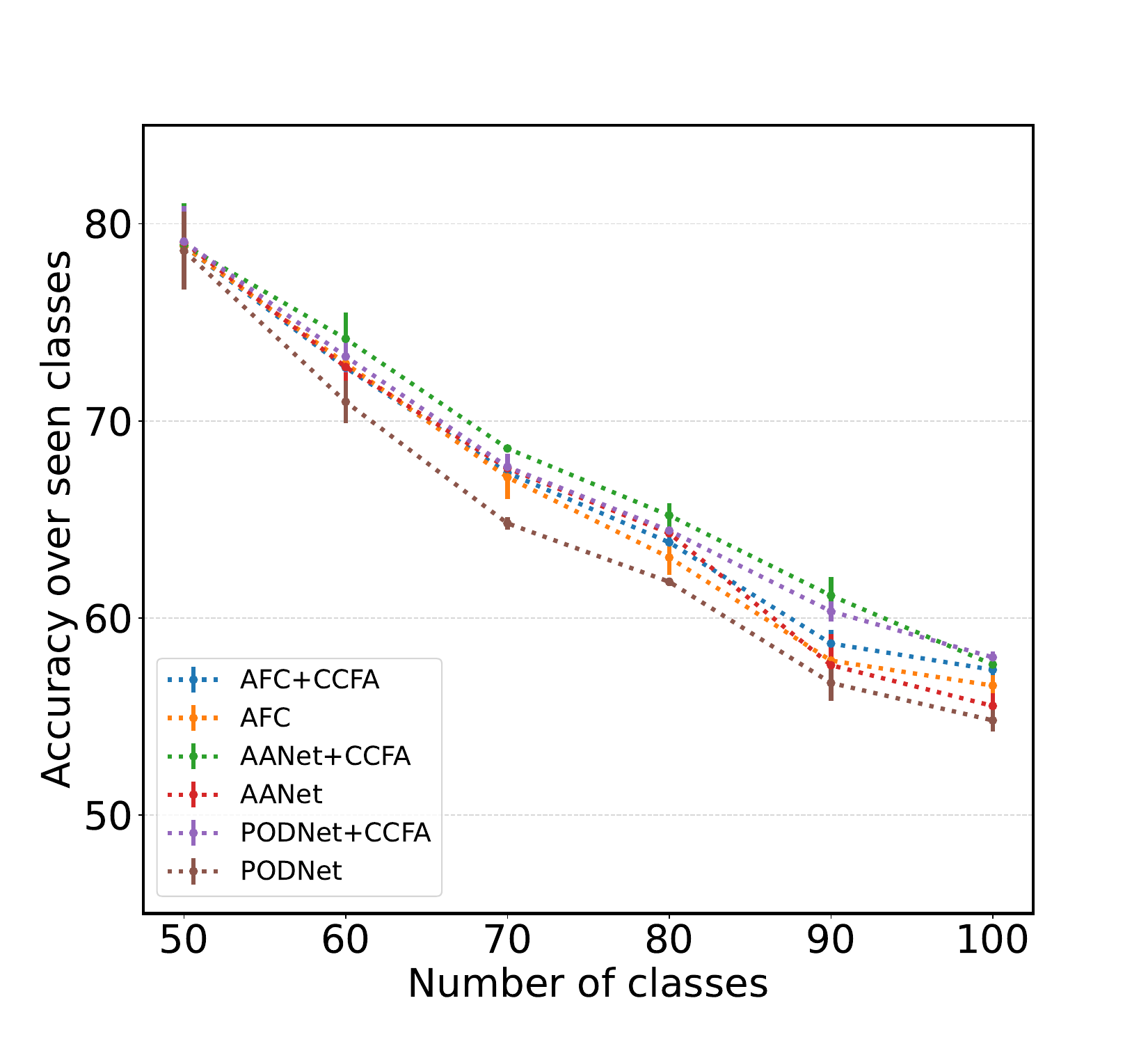}
          \vspace{-0.5cm}
         \caption{10$\times$5 stages}         
         \label{fig:cifar5step}
     \end{subfigure}
     \caption{Plots for accuracy on CIFAR-100 along with the incremental stages.}
     \label{fig:cifar}
\end{figure}

\section{Additional Results on CIFAR-100}
\label{sec:result_cifar}
We compare accuracy at each incremental stage on CIFAR-100 for UCIR~\cite{hou2019learning}, PODNet~\cite{douillard2020podnet}, AANet~\cite{liu2021adaptive} and AFC~\cite{kang2022class} with CCFA.
Figure~\ref{fig:cifar} presents the accuracy over seen classes at each incremental stages with varying task size.
From the plots, we can easily observe that the CCFA improves accuracy in most of the incremental stages on every task size. 
Table~\ref{tab:main_table_cifar} shows the average incremental accuracies of CCFA compared to baselines with standard deviation over multiple runs.

\begin{table*}[h!]
	\caption{Class incremental learning performance on CIFAR-100 for our model and state-of-the-art frameworks with standard eviation over multiple runs.
	The proposed algorithm consistently improves the performance when plugged into the existing methods.
	Models with $*$ are our reproduced results. 
	The bold-faced numbers indicate the best performance.}
	\label{tab:main_table_cifar}
	\vspace{-0.2cm}
	\begin{center}
		\scalebox{1.0}{
			\renewcommand{\arraystretch}{1.2}
			\setlength\tabcolsep{10pt}
			\begin{tabular}{lllll}
\multicolumn{1}{l}{Number of tasks}  & \multicolumn{1}{c}{50}      &  \multicolumn{1}{c}{25}   &  \multicolumn{1}{c}{10} &  \multicolumn{1}{c}{5}     \\ \midrule
				
				UCIR{*}~\cite{hou2019learning}                  &   50.44 $\pm$0.32  &  56.61 $\pm$0.32   & 60.47 $\pm$0.32         &  63.76 $\pm$0.32        \\
				 UCIR{*}~\cite{hou2019learning}   +  CCFA      &        \textbf{52.64} $\pm$1.17           &   \textbf{57.42} $\pm$0.70 & \textbf{61.58} $\pm$0.60         &  \textbf{64.13}$\pm$0.32    \\ \hline

				  PODNet{*}~\cite{douillard2020podnet}           &   57.84 $\pm$0.37           &   60.50 $\pm$0.62     &  62.77 $\pm$0.78     &  64.62 $\pm$0.65       \\ 
				 PODNet{*}~\cite{douillard2020podnet}   + CCFA  &  \textbf{60.69} $\pm$0.56 &   \textbf{62.91} $\pm$0.91 &  \textbf{65.50} $\pm$0.70   & \textbf{67.24} $\pm$0.70    \\ \hline
				 AANet{*}~\cite{liu2021adaptive}                   &  60.91$\pm$1.00  &   62.34 $\pm$1.17    &  64.49 $\pm$0.81      &  66.34 $\pm$ 0.76      \\ 
				 AANet{*}~\cite{liu2021adaptive}  +  CCFA      & \textbf{62.20} $\pm$1.31    &\textbf{63.74} $\pm$0.92      &  \textbf{66.16}$\pm$1.03     &  \textbf{67.37} $\pm$1.12  \\ \hline	 
				  AFC{*}~\cite{kang2022class}                   &  61.74 $\pm$0.68  &   63.78 $\pm$0.42     &  64.63 $\pm$0.51     &  66.02 $\pm$0.47      \\ 
				 AFC{*}~\cite{kang2022class}  +  CCFA      & \textbf{63.11} $\pm$0.18   &   \textbf{64.59}$\pm$0.41     &  \textbf{65.61} $\pm$0.57    &  \textbf{66.47} $\pm$0.62  \\ 	 
				 \bottomrule				 
			\end{tabular} 
		}
	\vspace{-0.5cm}	
	\end{center}
\end{table*}

\begin{figure}[h!]
     \centering
     \hspace*{-3mm}
     \begin{subfigure}[b]{0.45\linewidth}
         \centering
          \hspace*{-3mm}
           \includegraphics[width=1.1\linewidth]{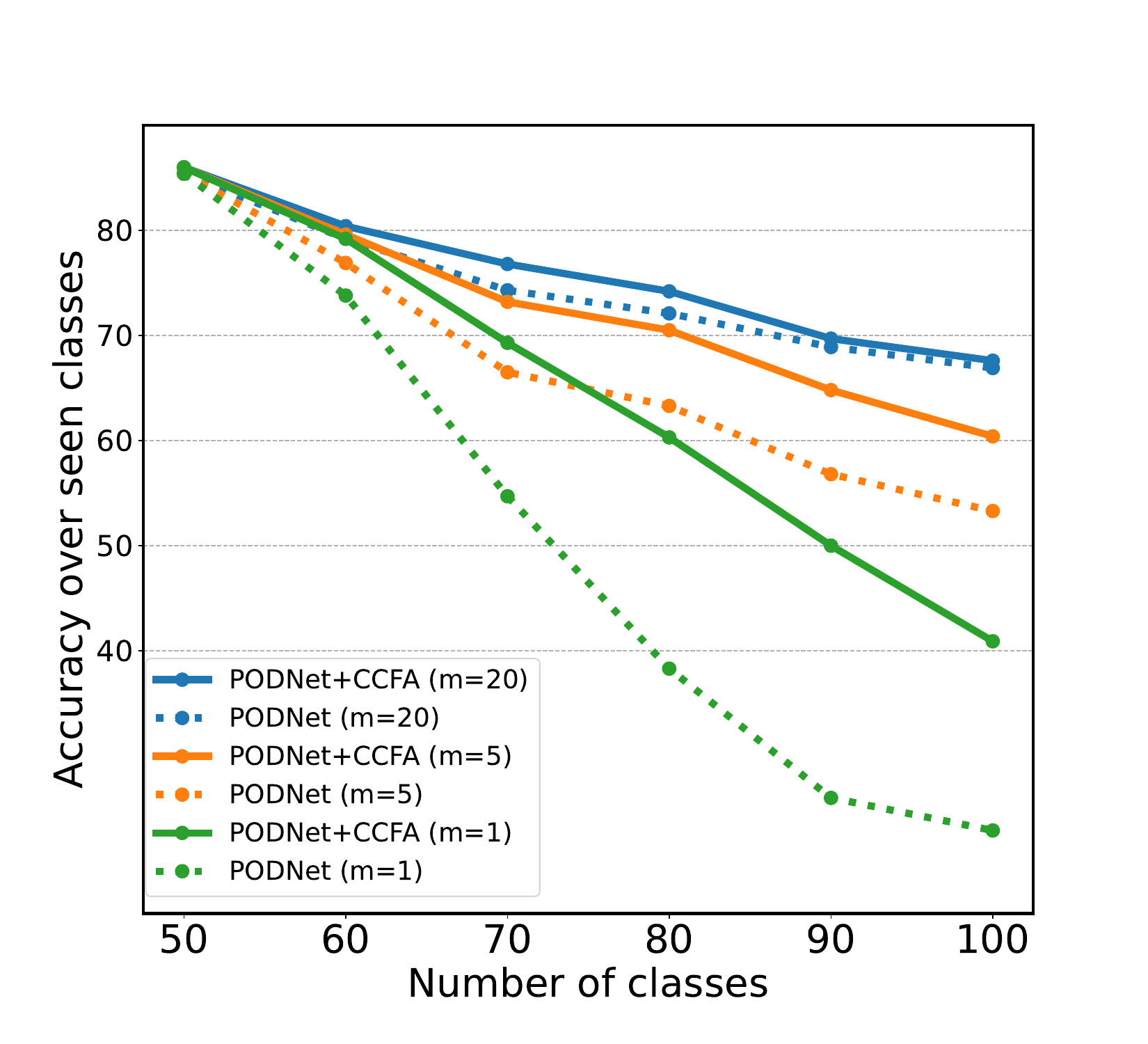}
           \vspace{-0.5cm}
          \caption{10$\times$5 stages}
         \label{fig:pod_10_100}
     \end{subfigure}
      \vspace{2mm}
     \begin{subfigure}[b]{0.45\linewidth}
         \centering
          \hspace*{-3mm}
           \includegraphics[width=1.1\linewidth]{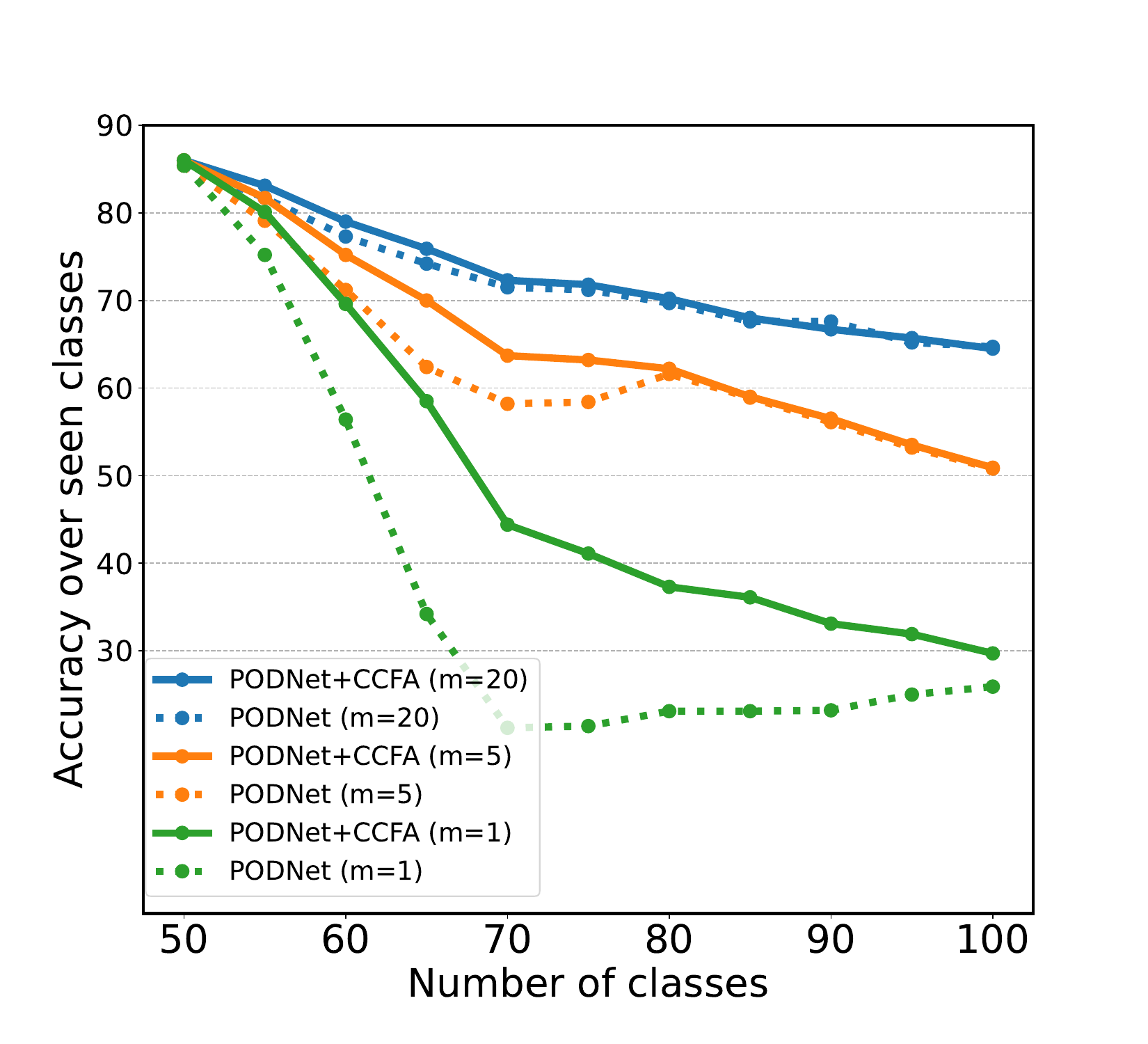}
           \vspace{-0.5cm}
          \caption{5$\times$10 stages}
         \label{fig:pod_10_100}
     \end{subfigure}
    
     \begin{subfigure}[b]{0.45\linewidth}
         \centering
          \hspace*{-3mm}
         \includegraphics[width=1.1\linewidth]{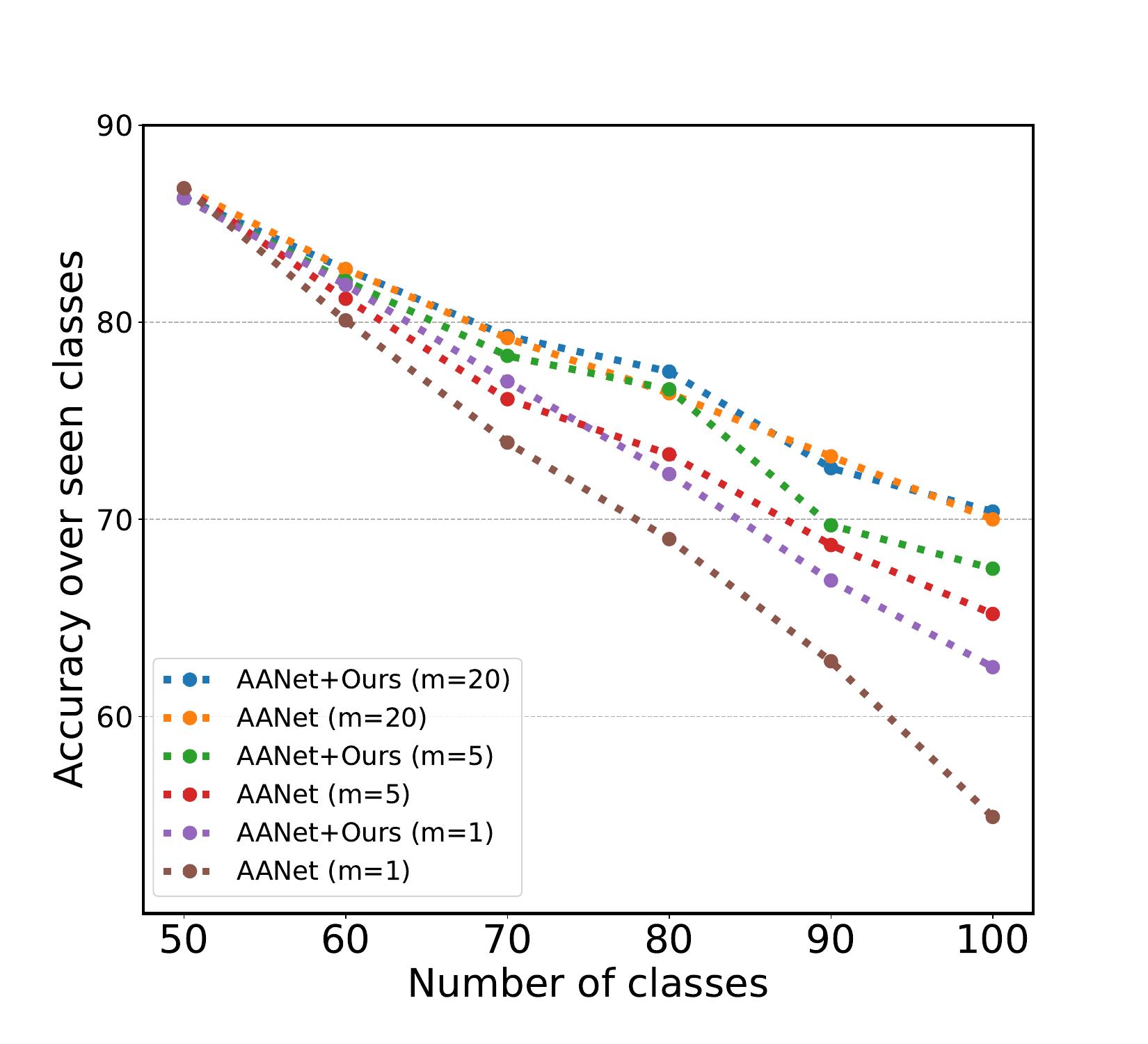}
          \vspace{-0.5cm}
         \caption{10$\times$5 stages}         
         \label{fig:aa_5_100}
     \end{subfigure}
 \vspace{2mm}
     \begin{subfigure}[b]{0.45\linewidth}
         \centering
          \hspace*{-3mm}
         \includegraphics[width=1.1\linewidth]{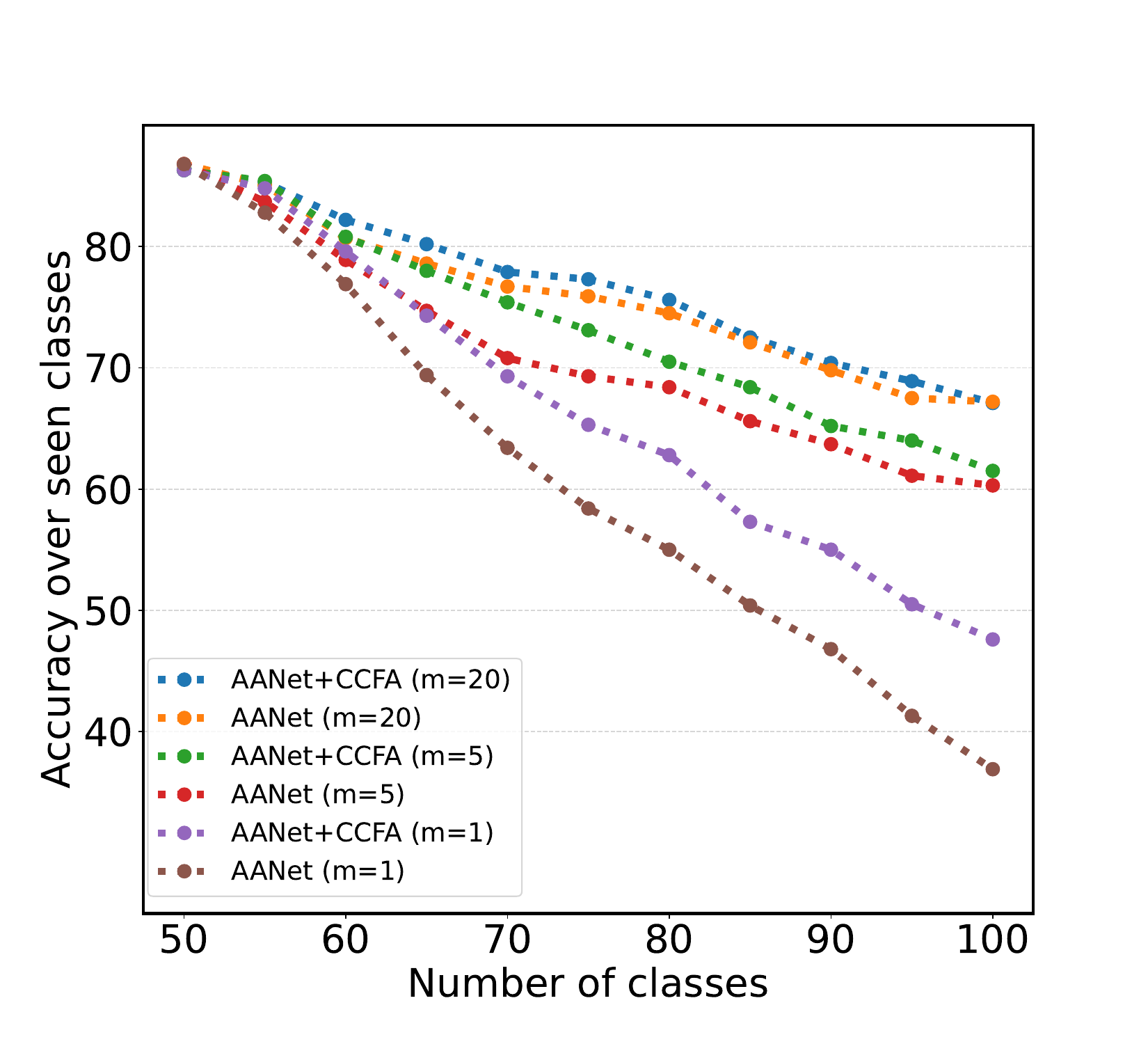}
          \vspace{-0.5cm}
         \caption{5$\times$10 stages}         
         \label{fig:aa_5_100}
     \end{subfigure}

     \begin{subfigure}[b]{0.45\linewidth}
         \centering
          \hspace*{-3mm}
         \includegraphics[width=1.1\linewidth]{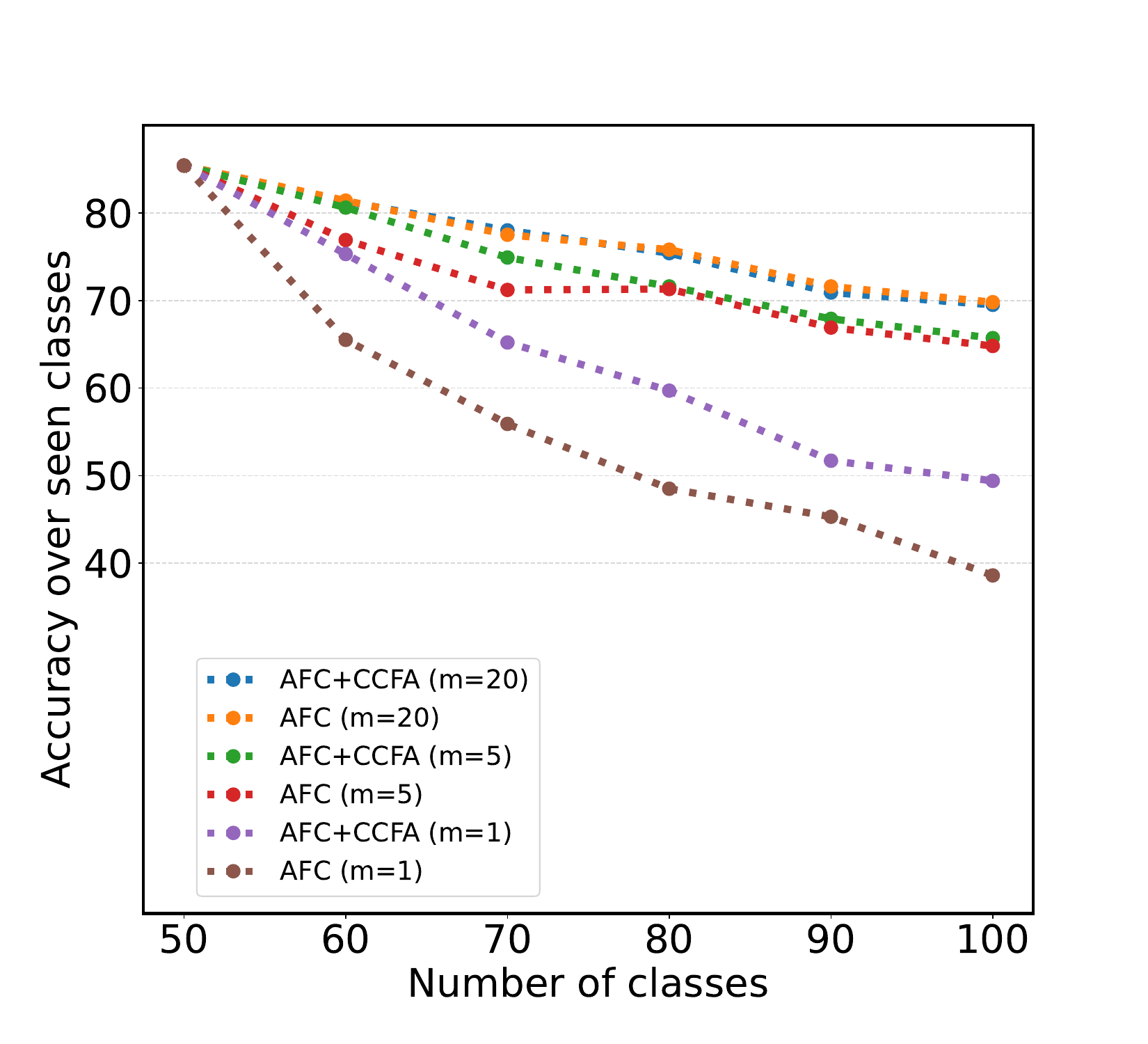}
          \vspace{-0.5cm}
         \caption{10$\times$5 stages}         
         \label{fig:afc_10_100}
     \end{subfigure}
  \vspace{2mm}
     \begin{subfigure}[b]{0.45\linewidth}
         \centering
          \hspace*{-3mm}
          \includegraphics[width=1.1\linewidth]{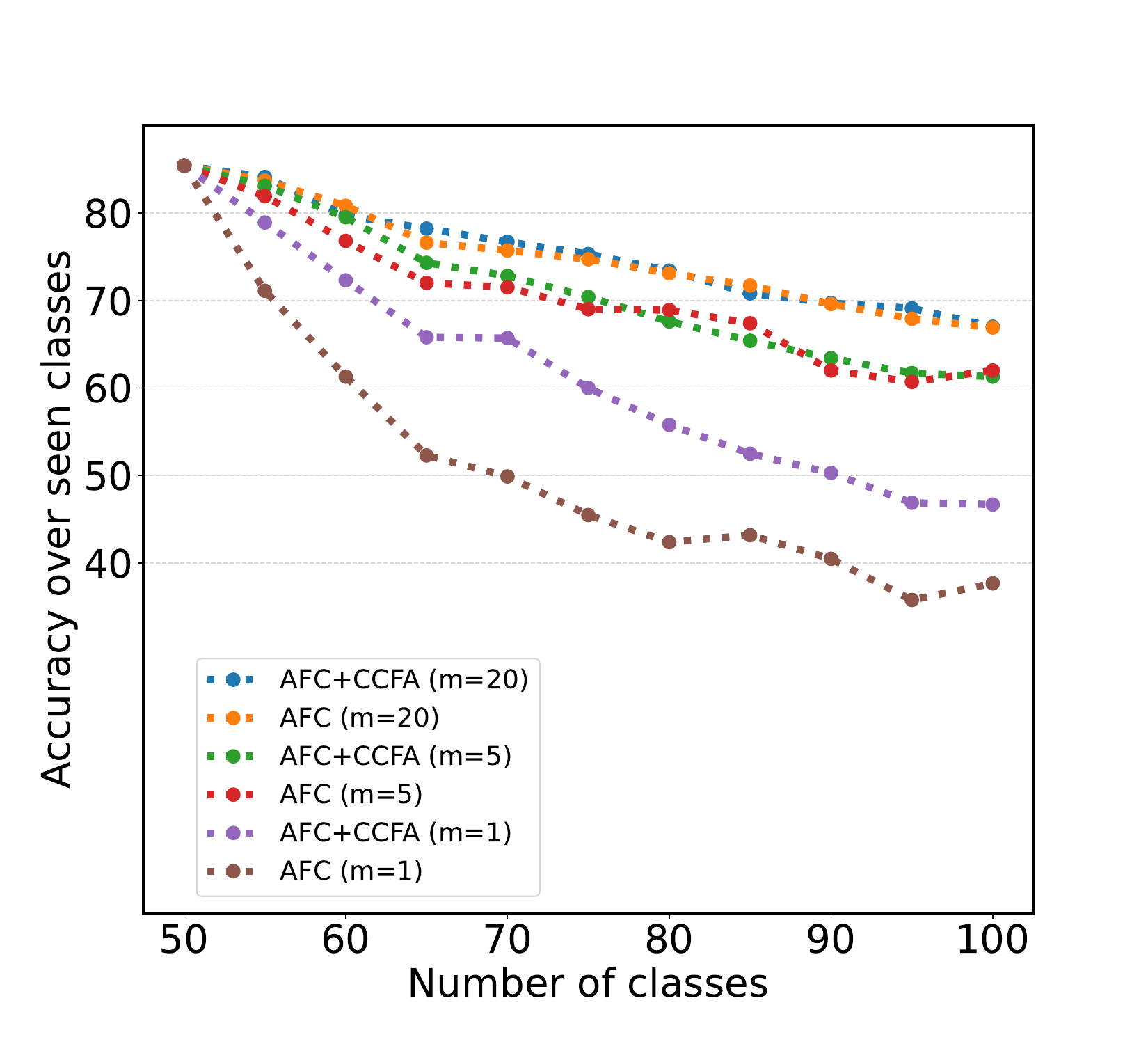}
          \vspace{-0.5cm}
         \caption{5$\times$10 stages}         
         \label{fig:afc_5_100}
     \end{subfigure}
     \caption{Plots for accuracy on ImageNet-100 along with the incremental stages.}
     \label{fig:imagenet100}
\end{figure}

\begin{figure}[t!]
     \centering
     \begin{subfigure}[b]{0.45\linewidth}
         \centering
          \hspace*{-3mm}
           \includegraphics[width=1.1\linewidth]{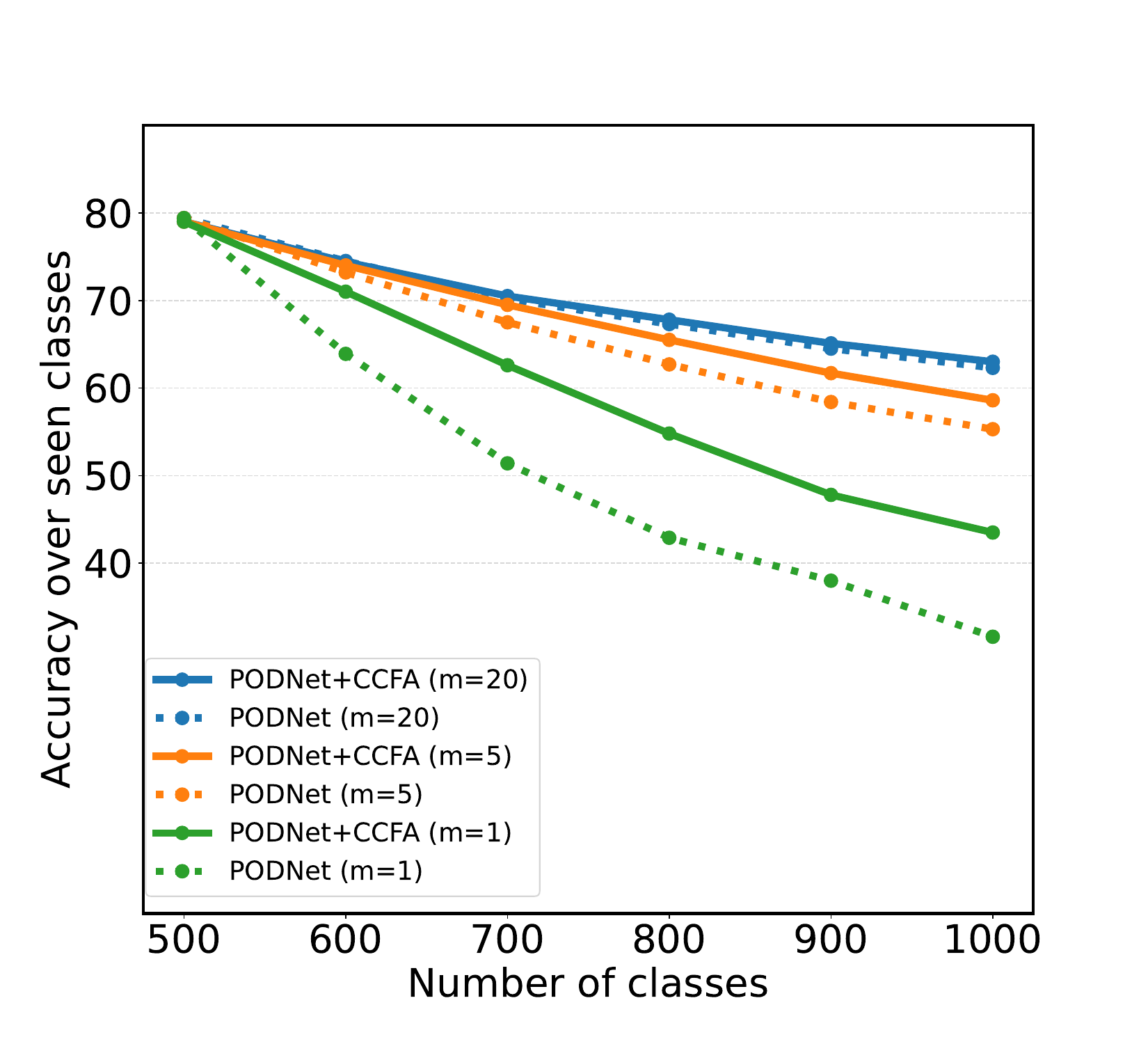}
           \vspace{-0.5cm}
          \caption{100$\times$5 stages}
         \label{fig:pod_10_100}
     \end{subfigure}
     \vspace{2mm}
     \begin{subfigure}[b]{0.45\linewidth}
         \centering
          \hspace*{-3mm}
           \includegraphics[width=1.1\linewidth]{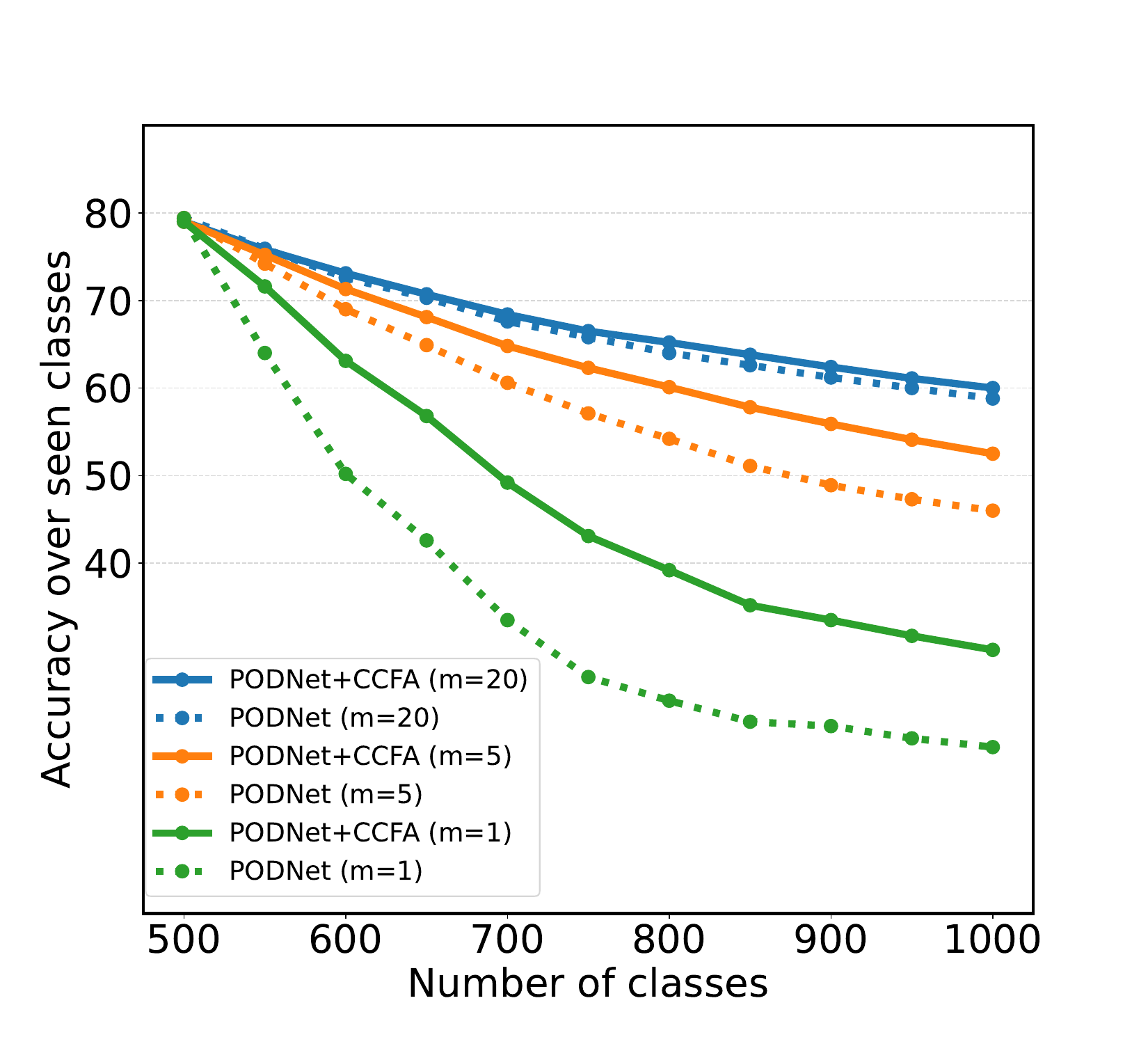}
           \vspace{-0.5cm}
          \caption{50$\times$10 stages}
         \label{fig:pod_10_100}
     \end{subfigure}

     \begin{subfigure}[b]{0.45\linewidth}
         \centering
          \hspace*{-3mm}
         \includegraphics[width=1.1\linewidth]{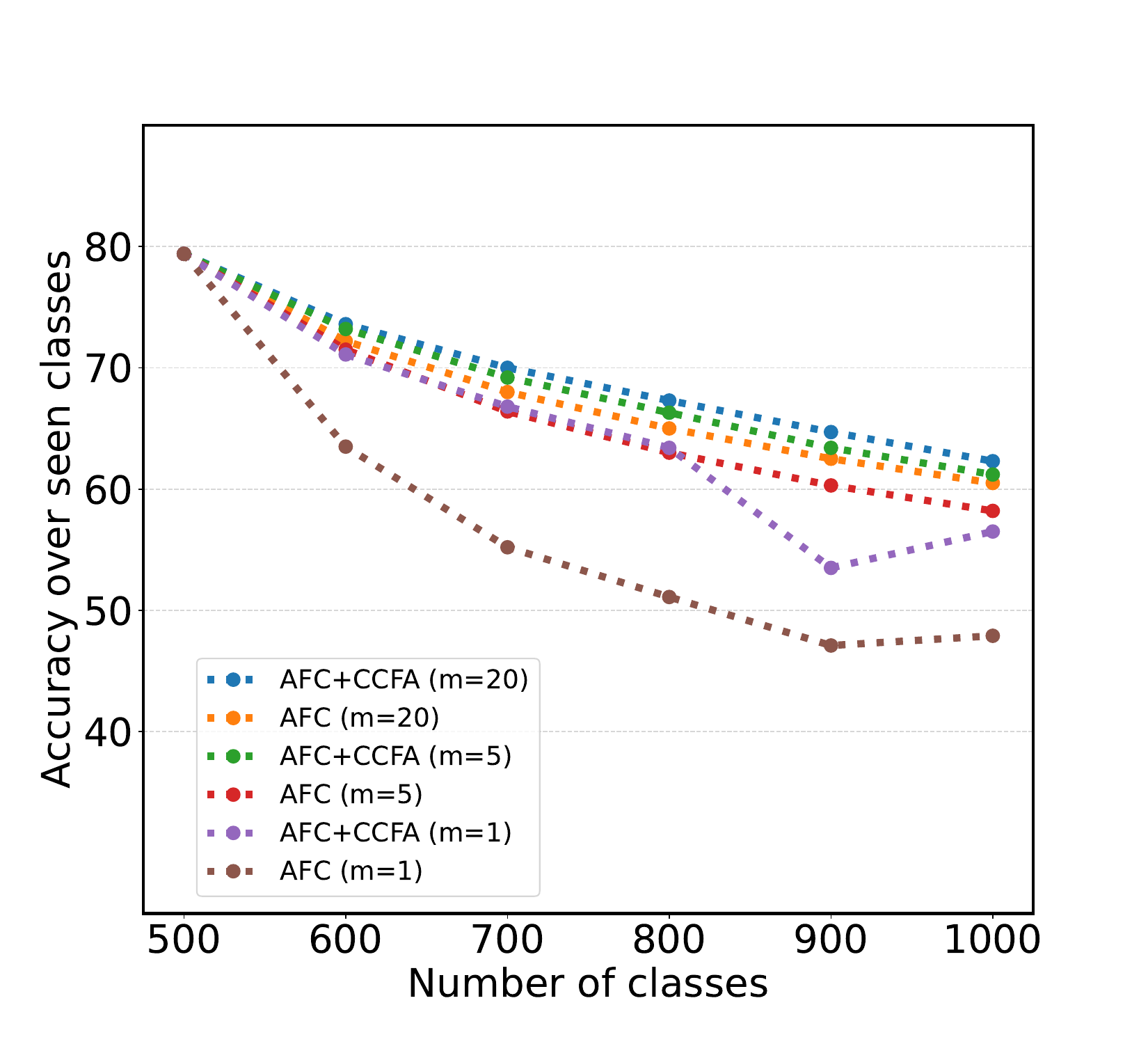}
          \vspace{-0.5cm}
         \caption{100$\times$5 stages}         
         \label{fig:aa_5_100}
     \end{subfigure}
     \vspace{2mm}
     \begin{subfigure}[b]{0.45\linewidth}
         \centering
          \hspace*{-3mm}
         \includegraphics[width=1.1\linewidth]{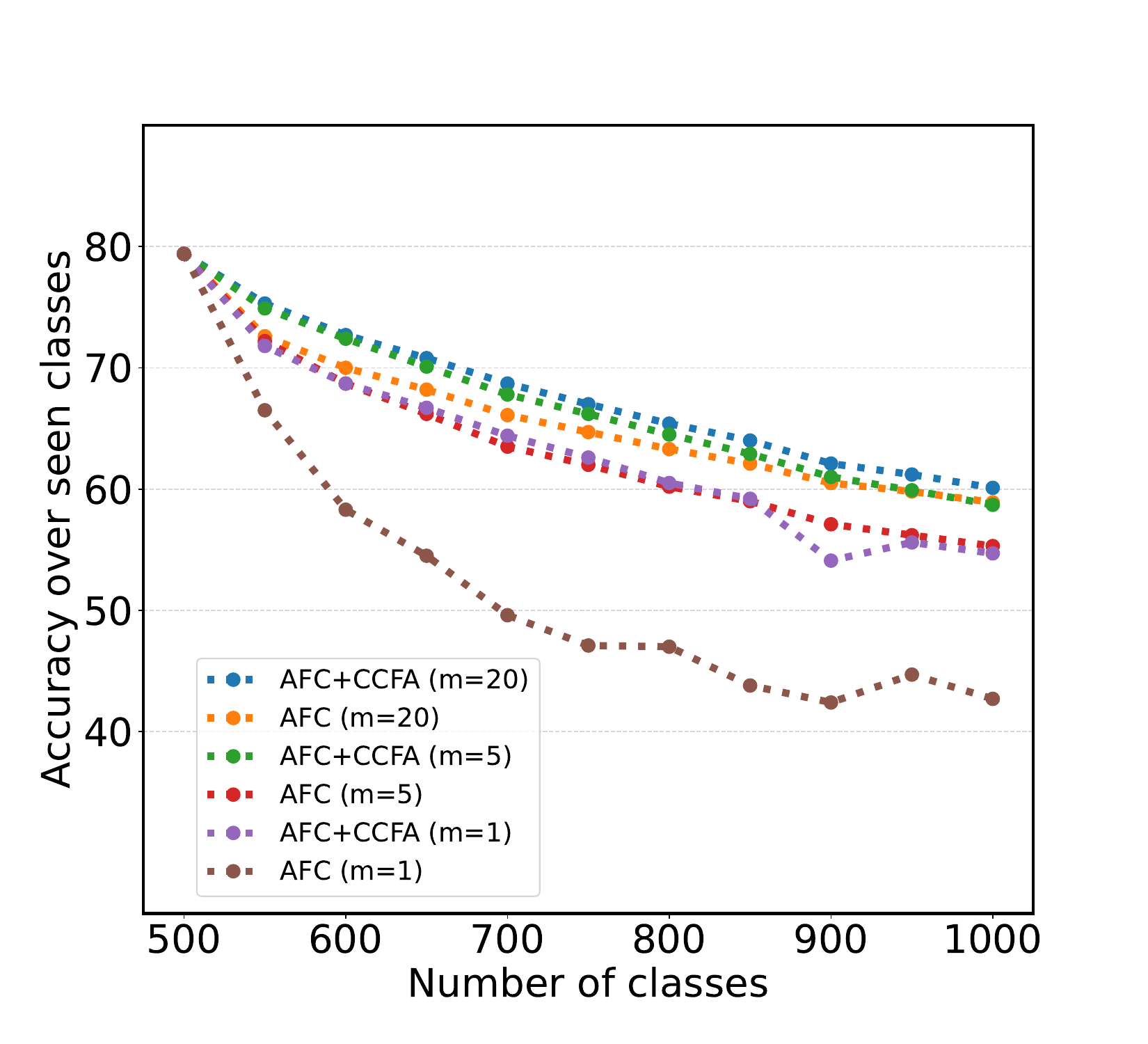}
          \vspace{-0.5cm}
         \caption{50$\times$10 stages}         
         \label{fig:afc_10_100}
     \end{subfigure}

      \caption{Plots for accuracy on ImageNet-1000 along with the incremental stages.}
     \label{fig:imagenet1000}
\end{figure}

\begin{algorithm}[t!]
  \caption{Cross-Class Feature Augmentation}
  \label{alg:ccfa}
  
  \begin{algorithmic}[1]
  \State{\bf Input:} A training example $(x,y)$, current feature extractor $f_{k}$, previous and current classifier, $g_{k-1}$ and $g_{k}$, number of iterations and step size for the PGD attack, $N$ and $\alpha$, old label set $\mathcal{Y}_{1:k-1}$
        \State $z_k = f_{k}(x)$ \Comment{Extract a feature} 
        \State $y_\text{target}$  $\gets \argmax_{\mathcal{Y}_{1:k-1}\setminus\{y \}} g_{k}({z_{k}})$ \Comment{Set the target class for the PGD attack} 
        \State $z_k' \gets z_k$ \Comment{Initialize the augmented feature}
         \For{$t = 0,\cdots,N-1$}
         \Comment{Apply the PGD attack} 
         \State $z'_{k} \gets z'_{k} - \alpha \sign (\nabla_{z} \mathcal{L}_\text{cls}(g_{k-1}(z'_k), y_\text{target}))$ 
         \EndFor
         \State $y_\text{ps} \gets \argmax_{\mathcal{Y}_{1:k-1}}g_{k-1}({z'_{k}}) $ \Comment{Get the pseudo-label for the augmented feature}
         \State{\bf Output:} Augmented feature representations $z'_{k}$, corresponding pseudo-label $y_\text{ps}$
  \end{algorithmic}
      
\end{algorithm}

\section{Additional Results on ImageNet-100 and ImageNet-1000}
\label{sec:result_imagenet}
We provide additional plots for accuracy on ImageNet-100 for PODNet~\cite{douillard2020podnet}, AANet~\cite{liu2021adaptive}, AFC~\cite{kang2022class} and ImageNet-1000 for PODNet~\cite{douillard2020podnet}, AFC~\cite{kang2022class} with CCFA at each incremental stage.
Figure~\ref{fig:imagenet100} and~\ref{fig:imagenet1000} show the performance of the baseline algorithms with CCFA on each incremental stage with different memory size and task size.
From the results, we conclude that CCFA boosts performance in most of the incremental stages in diverse settings.

\section{Algorithm Table for CCFA}
We provide the detailed algorithmic procedure of CCFA with a single example in Algoritnm~\ref{alg:ccfa}.
Note that target selection of CCFA is done by selecting the most confident classes for computational efficiency.
Even with this approximation, CCFA shows comparable results to results with target optimization processes.

\section{Implementation Details}
For the feature augmentation, we set the number of iterations for adversarial attack to 10 for all experiments.
Empirically, we find that $K=1$ is sufficient for top-K in target selection which is equivalent to selecting classes with highest confidence as $Y_{\text{target}}$ without solving LP optimization problem. 
In the CIFAR-100 experiments, we randomly sample the attack step size $\alpha$ from a uniform distribution $\mathcal{U}(\frac{2}{255},\frac{5}{255})$ and generate 640 features, which is 5 times the batch size, using 5 randomly sampled values of $\alpha$  for each real example.
For ImageNet, the attack step size $\alpha$ is randomly sampled from the same uniform distribution $\mathcal{U}(\frac{2}{2040},\frac{5}{2040})$ and 128 features, which is equal to the batch size, are generated using a random sample of $\alpha$.
We select the step size $\alpha$ in a way that the same amount of feature updates occurs in a single iteration for backbone networks with different feature dimensions.
For the number of augmented features, we consider the stability-plasticity trade-off.

\section{Details about the baseline algorithms}
We adopt UCIR~\cite{hou2019learning}, PODNet~\cite{douillard2020podnet}, AANet~\cite{liu2021adaptive} and AFC~\cite{kang2022class} as the baseline algorithms. 
Since CCFA is orthogonally applied to the existing class incremental learning algorithms, we briefly introduce the baseline algorithms in this section.
Moreover, we provide a detailed procedure of CCFA.
\subsection{UCIR}
UCIR~\cite{hou2019learning} introduces the cosine classifier employing cosine-normalized features and class embeddings to avoid the over-magnitude of the new classes due to the class imbalances. 
Instead of matching logits after the classification layer, UCIR matches the normalized features just before the classification layer as follows,
\begin{equation}\label{eq:lessforget}
\mathcal{L}_\text{dis} = 1-\langle f_{k-1}(x), f_{k}(x) \rangle,
\end{equation}
where $f_{k}(x)$ is the normalized feature from the feature extractor of $k^{\text{th}}$ incremental stage.
While the logit-based distillation loss can only force the angles between the features and old class embeddings to be fixed, $\mathcal{L}_\text{dis}$ in equation (\ref{eq:lessforget}) can fix the exact orientation of feature vectors at each incremental stage.

Another main component of UCIR is the margin-ranking loss, which tries to separate the old classes and new classes. 
For each exemplar sample $x$, UCIR finds top-$M$ nearest new embeddings and computes the margin-ranking loss, which is defined by
\begin{equation}\label{eq:marginranking}
\mathcal{L}_{\text{mr}} = \sum_{m=1}^{M} \max(\delta-\langle \theta(x), f_{k}(x) \rangle + \langle \theta_{m}(x), f_{k}(x)\rangle , 0) , 
\end{equation}
where $m$ is the margin threshold, $\theta(x)$ is  class embedding of ground truth class and $\theta_m(x)$ are class embeddings of top-$M$ nearest classes.
Integrated training objective of UCIR is defined by
\begin{equation} \label{eq:ucirobj}
\mathcal{L}_{final} = \mathcal{L}_{cls} + \lambda_{dis} \mathcal{L}_{dis} + \lambda_{mr} \mathcal{L}_{mr},
\end{equation}
where $\mathcal{L}_{cls}$ is standard cross-entropy loss and $\lambda_{dis}, \lambda_{mr}$ are weights for balancing the losses, respectively.

\subsection{PODNet}
PODNet~\cite{douillard2020podnet} tries to prevent forgetting by applying the Pooled Outputs Distillation (POD) which mimics, not only the final outputs of network, but also the intermediate activation maps. 
To provide certain degree of freedom for adaptivity to new classes, PODNet matches the pooled activation maps instead of directly matching the activation maps.
For the intermediate features, PODNet exploits two types of POD-losses, $\mathcal{L}_{\text{POD-width}}$ and $\mathcal{L}_{\text{POD-height}}$, which are given by
\begin{equation} \label{eq:podloss}
\begin{split}
\mathcal{L}_{\text{POD-width}}(h_{l}^{k-1}, h_{l}^{k}) & = \sum_{c=1}^{C} \sum_{h=1}^{H} \norm{\sum_{w=1}^{W} h_{l,c,w,h}^{k-1} - \sum_{w=1}^{W} h_{l,c,w,h}^{k}}, \\
\mathcal{L}_{\text{POD-height}}(h_{l}^{k-1}, h_{l}^{k}) & = \sum_{c=1}^{C} \sum_{w=1}^{W} \norm{\sum_{h=1}^{H} h_{l,c,w,h}^{k-1} - \sum_{h=1}^{H} h_{l,c,w,h}^{k}},
\end{split}
\end{equation}
where $h_{l}^{k}$ is the $l^{\text{th}}$ layer activation maps from feature extractor of $k^{\text{th}}$ incremental stage and $c, w, h$ stand for the channel, width, height dimension of the output activation.
Total distillation loss for intermediate features is defined by the sum of $\mathcal{L}_{\text{POD-width}}$ and $\mathcal{L}_{\text{POD-height}}$ as follows,
\begin{equation} \label{eq:podspatial}
\begin{split}
\mathcal{L}_{\text{POD-spatial}}(h_{l}^{k-1}, h_{l}^{k}) &= \\
\mathcal{L}_{\text{POD-width}}(h_{l}^{k-1}&, h_{l}^{k}) + \mathcal{L}_{\text{POD-height}}(h_{l}^{k-1}, h_{l}^{k}).
\end{split}
\end{equation}
PODNet also uses distillation loss for final outputs which is given by,
\begin{equation} \label{eq:PODflat}
\mathcal{L}_{\text{POD-flat}}(h^{k-1}, h^{k}) = \norm{h^{k-1} - h^{k}}^{2}.
\end{equation}
Final distillation loss of PODNet is defined by
\begin{equation} \label{eq:POD}
\begin{split}
\mathcal{L}_{\text{POD}} = \lambda_{s}& \sum_{l=1}^{L} \mathcal{L}_{\text{POD-spatial}}(f_{l}^{k-1}(x), f_{l}^{k}(x)) \\
&+ \lambda_{f} \mathcal{L}_{\text{POD-flat}}(f^{k-1}(x), f^{k}(x)), 
\end{split}
\end{equation}
where $\lambda_{s}, \lambda_{f}$ are weights for balancing two loss terms.
Instead of linear classifier, PODNet uses the ensemble classifier named local similarity classifier (LSC) to capture the finer local structures of decision boundaries. Logits of LSC $\hat{y}_{q}$ is computed as
\begin{equation} \label{eq:lsc}
s_{q,	p} = \dfrac{\text{exp} \langle \theta_{q,p}, h \rangle}{\sum_{i} \text{exp} \langle \theta_{q,i}, h \rangle}, \quad \hat{y}_{q} = \sum_{p}  s_{q,p} \langle \theta_{q,p}, h \rangle ,
\end{equation}
where $\theta_{q,p}$ stands for $p^{\text{th}}$ class embedding of class $q$.
With LSC, final classification loss $\mathcal{L}_{cls}$ is computed using NCA loss~\cite{movshovitz2017no} as follows, 
\begin{equation} \label{eq:lsc}
\mathcal{L}_{\text{cls}} = \left[ -\log \dfrac{\text{exp}(\eta (\hat{y}_{y}- \delta))}{\sum_{i \neq y} \text{exp } \eta \hat{y}_{i}} \right]_{+},
\end{equation}
where $y$ is ground-truth class, $\delta$ is a margin, $\eta$ is a temperature and $\left[ \cdot \right]_{+}$ is a hinge to bound a loss to be positive.
Integrated training objective of PODNet is defined by
\begin{equation} \label{eq:obj}
\mathcal{L}_{\text{final}} = \mathcal{L}_{\text{cls}} +  \mathcal{L}_{\text{POD}}.
\end{equation}

\subsection{AANet}
AANet~\cite{liu2021adaptive} is an orthogonal framework that can be applied to existing class incremental learning methods. 
AANet uses two homogeneous residual networks, which are named stable block and plastic block. 
Stable block aims to maintain the knowledge learned from old classes and plastic block fully adapts to samples from new classes.
For the stable block, AANet freezes the parameters $\theta_{base}$ learned in the first incremental stage and learns the small set of scaling weights $\phi_{k}$, which are applied to each neuron in $\theta_{base}$.
For the plastic block, the network updates all of its parameters $\theta_{k}$ as learning proceeds.
Given input $x^{l-1}$, AANet aggregates the outputs of plastic block and stable block at every residual level $l$ as follows
\begin{equation} \label{eq:aggregation}
\begin{split}
x_{\phi_{k}}^{l} &= f_{k}(x^{l-1}; \phi_{k}  \odot \theta_{\text{base}}), \quad x_{\theta_{k}}^{l} = f_{k}(x^{l-1}; \theta_{k}), \\
x^{l} &= \alpha_{\phi_{k}}^{l} \cdot x_{\phi_{k}}^{l} + \alpha_{\theta_k}^{l} \cdot x_{\theta_{k}}^{l} , 
\end{split}
\end{equation}
where $\alpha_{\phi_{k}}^{l}, \alpha_{\theta_k}^{l}$ are learnable aggregation weights.
Parameters $\phi_{k}, \theta_k$ and $\alpha_{\phi_{k}}, \alpha_{\theta_k}$ are optimized using Bilevel Optimization (BOP).
$\phi_{k}, \theta_k$ are learned using $\mathcal{D}'_{k}=\mathcal{D}_{k} \cup \mathcal{M}_{k-1}$.
After that, $\mathcal{M}_{k-1}$ is updated to $\mathcal{M}_{k}$ using herding strategies and $\alpha_{\phi_{k}}, \alpha_{\theta_k}$ are updated with $\mathcal{M}_{k}$.
Note that AANet adopts the loss terms from each baselines, UCIR and PODNet.

\subsection{AFC}
AFC~\cite{kang2022class} handles the catastrophic forgetting problem by minimizing the upper bound of the loss increases induced by representation change.
Assume that we have convolutional neural network $M^t(\cdot)$ at an incremental stage $t$, which consist of a classifier $g(\cdot)$ and $L$-layers building blocks $f_l(\cdot); l=1,2,\dots, L$.
For the feature maps $Z_l=\{ Z_{l,1}, \dots, Z_{l, C_{l}} \}$ from $l^{\text{th}}$ building block, AFC uses two sub-networks $F_l(\cdot)$ and $G_l(\cdot)$ that split the whole network as follows:
\begin{equation}
M^t(x)=G_l(Z_l)=G_l(F_l(x)).
\end{equation}
To resolve the catastrophic forgetting problem given by,
\begin{equation}
\underset{M_t}{\min} \: \mathbb{E}_{(x,y) \sim \mathcal{P}^{t-1}} [\mathcal{L}(M_t(x), y)],
\end{equation}
where $\mathcal{P}^{t-1}$ is the data distribution at $t-1^{\text{st}}$ incremental stage, AFC approximate the loss by first order Taylor approximation as follows:
\begin{equation}
\label{eq:loss}
\begin{split}
\mathcal{L}(G_l(Z'_{l}), y) &\approx \mathcal{L}(G_l(Z_{l}), y) \\ &
+ \sum_{c=1}^{C_l}\langle \nabla_{Z_{l,c}} \mathcal{L}  (G_l(Z_{l}), y), Z'_{l,c}-Z_{l,c} \rangle_F  , 
\end{split}
\end{equation}
where $Z'_l$ and $Z_l$ are obtained respectively by $M_t(\cdot)$ and $M_{t-1}(\cdot)$.
By Eq.~\ref{eq:loss}, loss increases incurred by the representation shift in the $c^{\text{th}}$ channel of $l^{\text{th}}$ layer can be defined as,
\begin{equation}
\Delta \mathcal{L}(Z'_{l,c}) \coloneqq \langle \nabla_{Z_{l,c}} \mathcal{L}  (G_l(Z_{l}), y), Z'_{l,c}-Z_{l,c} \rangle_F.
\end{equation}
AFC reduces the catastrophic forgetting by minimizing the surrogate loss function,
\begin{equation}
\label{eq:surr}
\underset{M_t}{\min} \: \sum_{l=1}^{L} \sum_{c=1}^{C_l} \mathbb{E}_{(x,y) \sim \mathcal{P}^{t-1}} [\Delta \mathcal{L}(Z'_{l,c}) ]^2
\end{equation}
To reduce the computational complexity induced by storing a large number of feature maps, AFC introduces an additional upper bound of Eq.~\ref{eq:surr} as follows:
\begin{equation}
\label{eq:deriv}
\begin{split}
&\mathbb{E}[ \Delta \mathcal{L}(Z'_{l,c})] \\
&= \mathbb{E}[\langle \nabla_{Z_{l,c}} \mathcal{L}  (G_l(Z_{l}), y), Z'_{l,c}-Z_{l,c} \rangle_F] \\ 
&\leq \mathbb{E} [\lVert \nabla_{Z_{l,c}} \mathcal{L}  (G_l(Z_{l}), y)  \rVert_F  \cdot \lVert  Z'_{l,c}-Z_{l,c} \rVert_F] \\
&\leq \sqrt{\mathbb{E}[\lVert \nabla_{Z_{l,c}} \mathcal{L}  (G_l(Z_{l}), y)  \rVert_F^2] \cdot \mathbb{E}[\lVert  Z'_{l,c}-Z_{l,c} \rVert_F^2]}.
\end{split}
\end{equation}
By Eq.~\ref{eq:deriv}, AFC introduces the practical objective function given by,
\begin{equation}
\label{eq:fi_loss}
\underset{M_t}{\min} \: \sum_{l=1}^{L} \sum_{c=1}^{C_l} I_{l,c}^{t}  \mathbb{E}_{x \sim \mathcal{P}^{t-1}} [\lVert Z'_{l,c}-Z_{l,c} \rVert^2],
\end{equation}
where the importance $I_{l,c}^{t}$ is defined as,
\begin{equation}
I_{l,c}^t \coloneqq  \mathbb{E}_{(x,y) \sim \mathcal{P}^{t-1}} [\lVert \nabla_{Z_{l,c}} \mathcal{L}  (G_l(Z_{l}), y)  \rVert_F^2].
\end{equation}
Due to the sparsity of previous task data, AFC minimizes the discrepancy loss $\mathcal{L}_{\text{disc}}$ corresponding to the upper bound of Eq.~\ref{eq:fi_loss} as follows:
\begin{equation}
\mathcal{L}_{\text{disc}} \coloneqq \sum_{l=1}^{L} \sum_{c=1}^{C_l} I_{l,c}^{t}  \mathbb{E}_{x \sim \mathcal{P}^{t}} [\lVert Z'_{l,c}-Z_{l,c} \rVert^2].
\end{equation}
The final loss of AFC at incremental stage $t$ is defined as below:
\begin{equation}
\mathcal{L}_{\text{final}}^{t} = \mathcal{L}_{\text{cls}} + \lambda_{\text{disc}} \cdot \lambda^{t}\mathcal{L}_{\text{disc}}^{t},
\end{equation}
where $\lambda_{\text{disc}}$ is a hyper-parameter and $\lambda^{t}$ is an adaptive weight for each incremental stage.

\bibliography{aaai24}